\documentclass[sigconf]{acmart}

\AtBeginDocument{%
  }

\setcopyright{acmlicensed}
\copyrightyear{2025}
\acmYear{2025}
\acmDOI{XXXXXXX.XXXXXXX}

\acmConference[WWW '25]{In Proceedings of the ACM Web Conference 2025}{April 28-May 02, 2025}{Sydney, Australia}
\usepackage{subfigure}
\usepackage{listings}
\usepackage{algorithm,algcompatible}
\algnewcommand\algorithmicreturn{\textbf{return}}
\algnewcommand\RETURN{\algorithmicreturn}

\usepackage{multirow} 
\usepackage{enumitem} 
\usepackage{amsmath}
\usepackage{xcolor}
\usepackage{subfigure}
\usepackage{mwe}
\usepackage{booktabs}
\usepackage{threeparttable}
\usepackage{amsmath}
\usepackage{amsfonts}
\usepackage[switch]{lineno}
\usepackage{flushend}

\begin{document}

\title{Unveiling Discrete Clues: Superior Healthcare Predictions for Rare Diseases}

\author{Chuang Zhao}
\affiliation{%
  \institution{The Hong Kong University of Science and Technology }
  \city{Hong Kong}
  \country{China}}
  \email{czhaobo@connect.ust.hk}
  
  \author{Hui Tang}
\affiliation{%
  \institution{The Hong Kong University of Science and Technology }
  \city{Hong Kong}
  \country{China}}
  \email{eehtang@ust.hk}

  \author{Jiheng Zhang}
\affiliation{%
  \institution{The Hong Kong University of Science and Technology }
  \city{Hong Kong}
  \country{China}}
  \email{jiheng@ust.hk}

  \author{Xiaomeng Li}
  \authornote{Xiaomeng Li is the corresponding author.}
\affiliation{%
  \institution{The Hong Kong University of Science and Technology }
  \city{Hong Kong}
  \country{China}}
  \email{eexmli@ust.hk}
  
\renewcommand{\shortauthors}{Chuang Zhao et al.}

\begin{abstract}
Accurate healthcare prediction is essential for improving patient outcomes. 
Existing work primarily leverages advanced frameworks like attention or graph networks to capture the intricate collaborative (CO) signals in electronic health records.
However, prediction for rare diseases remains challenging due to limited co-occurrence and inadequately tailored approaches.
To address this issue, this paper proposes UDC, a novel method that \underline{u}nveils  \underline{d}iscrete \underline{c}lues to bridge consistent textual knowledge and CO signals within a unified semantic space, thereby enriching the representation semantics of rare diseases.
Specifically, we focus on addressing two key sub-problems: (1) acquiring distinguishable discrete encodings for precise disease representation and (2) achieving semantic alignment between textual knowledge and the CO signals at the code level.
For the first sub-problem, we refine the standard vector quantized process to include condition awareness. Additionally, we develop an advanced contrastive approach in the decoding stage, leveraging synthetic and mixed-domain targets as hard negatives to enrich the perceptibility of the reconstructed representation for downstream tasks.
For the second sub-problem, we introduce a novel codebook update strategy using co-teacher distillation. This approach facilitates bidirectional supervision between textual knowledge and CO signals, thereby aligning semantically equivalent information in a shared discrete latent space.
Extensive experiments on three datasets demonstrate our superiority.
\end{abstract}



%

\maketitle

\section{Introduction}\label{sec:intro}
Healthcare predictions, such as medication recommendations, are critically important as they directly influence the efficacy of medical treatments~\cite{goh2021artificial,lv2024boxcare}. Accurate medication recommendations can enhance patient recovery rates by up to 30\% and reduce adverse drug reactions by 25\%, demonstrating their significant positive impact~\cite{vizirianakis2011nanomedicine,sadee2023pharmacogenomics}. 

\begin{figure}[!t]
  \centering
\setlength{\abovecaptionskip}{-0.05cm}   
\setlength{\belowcaptionskip}{-0.1cm}   
\subfigure[Occurence Distribution]{
\begin{minipage}[t]{0.48\linewidth}
\centering
\includegraphics[width=\linewidth,height=0.7\linewidth]{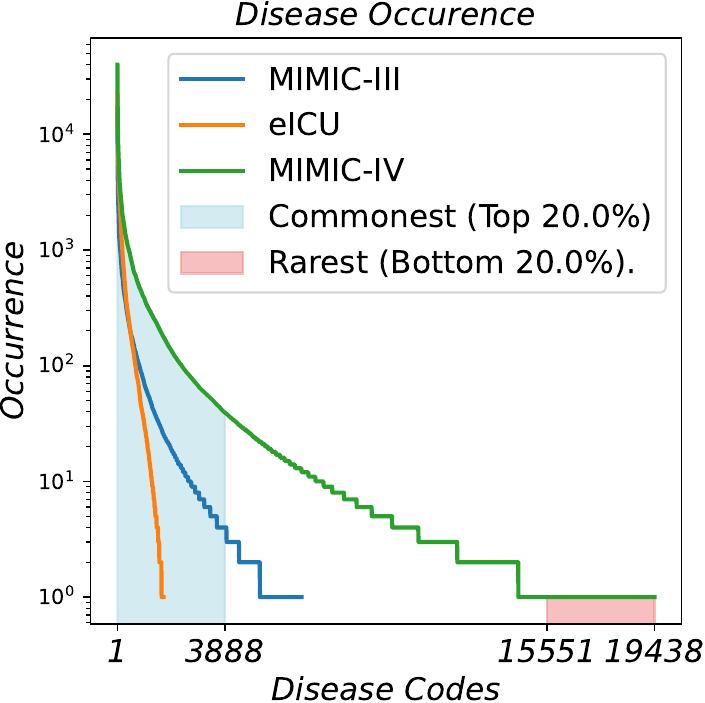}
\label{fig:motiv:dist}
\end{minipage}%
}%
\subfigure[Med Rec (MIMIC-IV)]{
\begin{minipage}[t]{0.48\linewidth}
\centering
\includegraphics[width=\linewidth,height=0.7\linewidth]{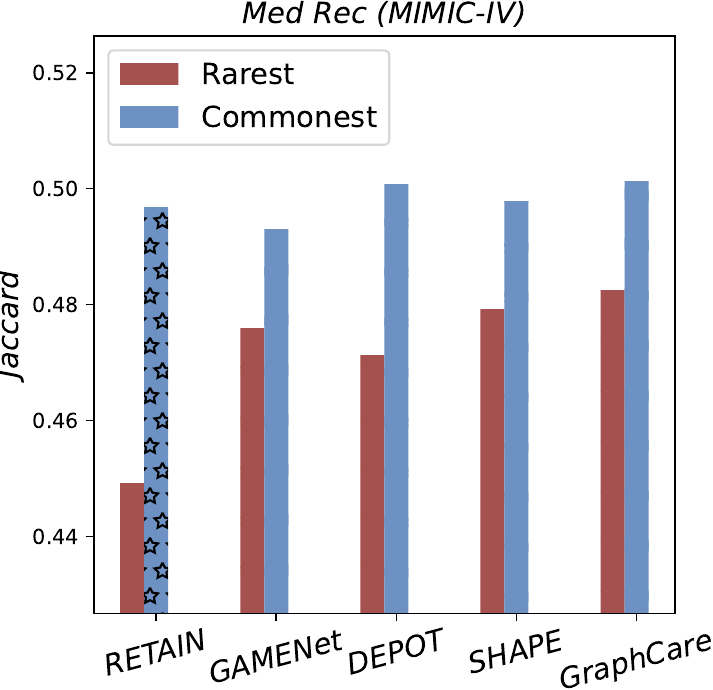}
\label{fig:motiv:per}
\end{minipage}%
}%
   \caption{(a) Disease occurrences across three datasets. (b) Medication recommendation for commonest / rarest diseases.}
   \label{fig:motiv}
   \vspace{-1em}
\end{figure}

Current research in healthcare prediction can be broadly categorized into three genres~\cite{ali2023deep,yang2023pyhealth,ma2020rare}: rule-based, graph-based, and sequence-based approaches. Rule-based systems~\cite{de2022guidelines,shang2019gamenet} typically rely on expert-defined rules to guide predictions, offering effective solutions but often facing limitations in scalability and potential conflicts among rules. In contrast, graph-based methods~\cite{choi2017gram,bhoi2021personalizing} leverage graph neural networks to model electronic health records (EHRs) as homogeneous or heterogeneous graphs, enhancing predictive performance through the exploration of intricate collaborative (CO) signals within the data. Sequence-based methods~\cite{zhao2024enhancing,yang2021change} represent a shift from static approaches by focusing on the sequential patterns inherent in longitudinal EHRs, capturing temporal dependencies that static models might overlook. While these methods are effective, they tend to emphasize maximizing overall accuracy~\cite{zhao2024leave,xu2023seqcare}, which can lead to performance degradation for specific diseases. This issue arises from the highly skewed data distribution in EHRs. As depicted in Figure~\ref{fig:motiv:dist}, datasets such as MIMIC-III~\cite{johnson2016mimic}, MIMIC-IV~\cite{johnson2020mimic}, and eICU~\cite{pollard2018eicu} exhibit a pronounced imbalance in data distribution. In MIMIC-IV dataset, the commonest diseases (top 20\%) account for approximately 95\% of interactions in EHRs, while the rarest diseases (tail 20\%) represent only about 0.2\%.
Meanwhile, as shown in Figure~\ref{fig:motiv:per}, we observe that existing advanced methods demonstrate superior performance in diagnosing common diseases. However, their effectiveness diminishes significantly when applied to rare diseases.
This disparity is a key factor contributing to overall predictive shortcomings and may lead to health inequalities in diagnosis~\cite{zhao2024leave}. It underscores the need for more effective strategies.

Recently, several studies have demonstrated distinct distributions of long-tail and head objects~\cite{zhang2023deep}. 
This observation motivates us to treat rare diseases and common diseases as different feature domains and find a way to align rare diseases (CO space) with common diseases (CO space) to leverage the established knowledge, e.g., disease-medication relationships derived from rich EHRs associated with common diseases. However, as depicted in Figure~\ref{fig:motiv}, limited data impedes the establishment of a robust CO space for rare diseases. 
Textual knowledge (Text), shared across all diseases and recognized as a consistent and reliable semantic resource~\cite{kim2023melt,wu2024supporting}, serves as a bridge to facilitate alignment between these two spaces.
Consequently, our aim is to align CO signals with textual knowledge within a unified discrete space, followed by executing a high-quality Text$\rightarrow$CO mapping for rare diseases to enrich representation semantics. 
The discrete space, derived from VQ-VAE~\cite{razavi2019generating}, employs a vector quantized (VQ) process to facilitate code-level mappings between textual knowledge and CO signals. This aligns with the multi-symptom nature of the disease and demands fewer computations compared to continuous modeling~\cite{ekerljung2011multi,lee2022autoregressive}.
To develop our approach, we highlight two key aspects. 
\begin{itemize}[leftmargin=12pt]
\item \textbf{How to acquire distinguishable discrete encodings for precise disease representation?} 1) In clinical documentation, even minor variations in symptoms can necessitate different medical codes, despite similar text descriptions. For instance, {Type 1} and {Type 2} diabetes, though both may present as "diabetes without complications," diverge significantly in their pathophysiology and management, with {Type 1} typically requiring lifelong insulin therapy and {Type 2} often managed through lifestyle modifications and oral medications. This necessitates that the model be adept at discerning subtle yet significant differences in clinical context, despite relatively similar text descriptions. 

2) While VQ-VAE is effective at reconstructing data and learning broad patterns, its approach to feature extraction and reconstruction may not always align with the specific, detailed requirements of downstream predictive tasks, resulting in potential limitations in predictive accuracy.
For example, while the reconstructed text representation provides a coherent overview, it might lack critical details like specific symptom patterns or treatment adherence levels. Similarly, reconstructed CO signals might miss key interactions or subtle patterns that are crucial for precise medication recommendation or diagnosis prediction.

\item \textbf{How to perform effective semantic alignment between CO signals and textual knowledge?} 
Text and CO signals typically reside in distinct semantic spaces, with text represented in natural languages and CO signals in interaction embeddings. This domain gap is an obstacle that hinders the Text$\rightarrow$CO signal mapping.  Furthermore, as both representations of disease are mapped into a discrete space—where each code embodies unique symptom semantics—aligning at the code level is crucial for mitigating the domain gap and facilitating knowledge transfer.
\end{itemize}

To tackle these challenges, we introduce UDC, a tailored VQ-VAE framework for healthcare that utilizes textual knowledge and CO signals for alignment and reconstruction, enhancing the representation semantics of rare diseases during discrete representation learning (DRL).
To ensure the distinguishability of disease encodings, we upgrade the original VQ process to incorporate condition-aware calibration. We specifically include medical entities that co-occur during the same visit for a particular disease as contextual conditions. This adjustment allows the model to produce distinct reconstructions based on varying contexts, even when the text appears similar. For instance, in a medical scenario, the distinction between Type 1 and Type 2 diabetes could be identified by examining complications such as diabetic ketoacidosis (more common in Type 1) or by specific laboratory findings in EHRs, thereby enhancing the granularity of representations.
Furthermore, to guarantee task relevance in the reconstructed representations, we devise a contrastive task-aware calibration. Leveraging mixed-domain and synthetic target representations as hard negatives, we boost the model's ability to discern distinct features and facilitate the reciprocal transfer of knowledge between CO signals and textual information. This empowers the reconstructed representations to react adaptively in accordance with the particular downstream tasks at hand.
To achieve better semantic alignment of Text-CO signals, we introduce a novel codebook update strategy using co-teacher distillation. In this approach, the text and the CO signal, both featuring encoded diseases, act as mutual reconstruction labels, facilitating the aggregation of quantized vectors encoded from two signals with equivalent semantics into a unified latent space.

To sum up, our key contributions are as follows.
\begin{itemize}[leftmargin=12pt]
    \item To our knowledge, UDC has significantly enriched the semantics of rare diseases, thereby improving healthcare prediction performance. Our framework can be seamlessly integrated into various advanced healthcare prediction models.
    \item We tailor the VQ process for healthcare, incorporate condition-aware and task-aware calibration, and devise a novel codebook update mechanism. These enhancements notably improve  reconstruction performance and adaptability to downstream tasks.
    \item Our algorithm demonstrates superior performance across two healthcare prediction tasks on three datasets, effectively handling both common and rare diseases. We have made the code available on Github~\footnote{https://github.com/Data-Designer/UDCHealth/README.md} to ensure reproducibility.
\end{itemize}

\section{Related Work}\label{sec:rela}
We review related work, emphasizing connections and distinctions.


\subsection{Healthcare Prediction}\label{sec:2.1}
Healthcare prediction employs advanced data-driven models to forecast clinical outcomes and disease progression~\cite{yang2023pyhealth}. This practice significantly impacts personalized treatment by facilitating early intervention and optimizing clinical decisions.


The primary genres in healthcare prediction include rule-based, graph-based, and sequence-based models. Rule-based models~\cite{shang2019gamenet,de2022guidelines}, stemming from clinical expertise, offer interpretability and ease of implementation. However, their limitations lie in adapting to dynamic patient data and conflict rules, hindering their efficacy. In contrast, graph-based models diverge as they are entirely data-driven. They intricately map relationships among clinical entities as nodes and edges within a graph framework~\cite{choi2017gram,bhoi2021personalizing}, excelling in modeling relational data and uncovering hidden patterns. However, they can be computationally intensive and encounter scalability challenges when applied to extensive datasets. On the other hand, sequence-based models~\cite{gao2020stagenet,liu2023SHAPE}, leveraging temporal data like longitudinal EHRs, dynamically capture temporal dependencies. This paradigm is typically constructed using architectures such as RNNs and Transformers. When combined with medical prior knowledge, it effectively captures the patient's condition. 
Recently, hybrid models~\cite{xu2023seqcare,jianggraphcare} have been introduced, combining these genres to harness their respective strengths. 
A common approach involves representing visit-level data as subgraphs or introducing external knowledge, followed by information extraction to incorporate both temporal and high-order CO signals.
While effective, most methods primarily aim to enhance overall accuracy, with limited focus on the unique challenges associated with the sparse rare diseases.

\textit{Our approach operates within this hybrid genre, specifically targeting the enhancement of rare disease prediction through the integration of textual knowledge. Leveraging discrete learning, our method effectively bridges textual knowledge with CO signals, bolstering the representation semantics tailored to rare diseases.} 


\subsection{Generative Retrieval}
Generative retrieval is a key technique in modern systems, enabling the direct generation of candidate items rather than selecting from a fixed set, as in discriminative genres~\cite{kuo2024survey}. This is critical for delivering context-aware retrievals in domains with limited data. 

Generative retrieval~\cite{sun2024learning,li2024matching} can be broadly categorized into three  genres: autoregressive-based~\cite{xiao2023survey,li2023generative}, GAN-based~\cite{chakraborty2024ten,jin2023diffusionret}, and autoencoder-based models~\cite{berahmand2024autoencoders,xia2024achieving,zheng2022movq}. Autoregressive models~\cite{li2023generative}, such as those utilizing Transformer architectures, generate sequences by predicting the next item based on previous context, making them well-suited for tasks requiring a sequential understanding. However, they are often computationally intensive and may suffer from exposure bias. GAN-based models~\cite{chakraborty2024ten} generate realistic candidate items through a generator that creates samples and a discriminator that evaluates their authenticity. While GANs~\cite{jin2023diffusionret} excel in producing high-quality outputs, they are challenging to train and may experience instability issues. Autoencoder-based models, including approaches like VAE~\cite{zhang2019d,rajput2024recommender}, use an encoder to map inputs to a latent space and a decoder to reconstruct them. These models effectively capture complex data distributions and facilitate structured, interpretable generation. VQ-VAE~\cite{razavi2019generating}, in particular, leverages discrete latent variables, balancing the strengths of both autoregressive and autoencoder-based approaches while offering robustness in handling diverse distributions.

\textit{
Our method aligns with the last genre, specifically extending VQ-VAE to healthcare. We focus on enhancing the representation of rare diseases by introducing condition-aware and task-aware calibration.
Furthermore, we devise a novel co-teacher distillation to achieve code-level semantic alignment.
These tailored advancements enhance the accuracy and relevance of the rare disease representations generated, thereby boosting the performance of VQ-VAE within  healthcare tasks. 
}

\section{Proposed Method}\label{sec:method}

\noindent\textbf{Preliminary.}
Each patient's medical history is recorded as a sequence of visits, represented by $\mathcal{U}^{(k)} = (\mathbf{u}^{(k)}_{1}, \mathbf{u}^{(k)}_{2}, \dots, \mathbf{u}^{(k)}_{\mathcal{T}_{k}})$, where $k$ identifies the patient within the patient set $\mathcal{N}$, and $\mathcal{T}_{k}$ is the total number of visits. Each visit $\mathbf{u}^{(k)}_{t}$ is defined as a triplet $\mathbf{u}^{(k)}_{t} = (\mathbf{d}^{(k)}_{t}, \mathbf{p}^{(k)}_{t}, \mathbf{m}^{(k)}_{t})$, corresponding to the diagnoses ($d$), procedures ($p$), and medications ($m$) associated with that visit, respectively. These components are encoded as multi-hot vectors: $\mathbf{d}^{(k)}_{t} \in \{0,1\}^{|\mathcal{D}|}$, $\mathbf{p}^{(k)}_{t} \in \{0,1\}^{|\mathcal{P}|}$, and $\mathbf{m}^{(k)}_{t} \in \{0,1\}^{|\mathcal{M}|}$, where $\mathcal{D}$, $\mathcal{P}$, and $\mathcal{M}$ represent the sets of all possible diagnoses, procedures, and medications, and $|\cdot|$ denotes the cardinality of these sets. For instance, the vector $\mathbf{d} = [1,0,1,0]$ suggests that the patient has diseases 1 and 3, assuming $|\mathcal{D}|=4$.  Additionally, each medical entity $*$ is associated with a corresponding text description denoted as $\mathrm{T}(*)$.
For clarity, $k$ is omitted in the following content.

\noindent\textbf{Task formulation.} Following~\cite{zhao2024enhancing,ye2021medpath,yang2023molerec}, we outline the definitions of the two common healthcare prediction tasks. 

\begin{itemize}[leftmargin=12pt]
\vspace{-0.3em}
 \item \textbf{Diagnosis Prediction (Diag Pred)} entails a multi-label classification challenge that centers on anticipating forthcoming risks. This task revolves around scrutinizing $[\mathbf{u}_{1}, ..., \mathbf{u}_{t}]$ to forecast the diagnosis set $\mathbf{d}_{t+1}$ at time $t+1$, where target $\mathbf{y}[\mathbf{u}_{t+1}] \in \mathbb{R}^{1\times |\mathcal{D}|}$.

 \item \textbf{Medication Recommendation (Med Rec)} involves a multi-label classification task dedicated to pinpointing the most suitable medications for the patient's present state. This process entails scrutinizing $[{\mathbf{u}_{1}, ..., \mathbf{u}_{t}}]$, alongside $(\mathbf{d}_{t+1}, \mathbf{p}_{t+1})$, to anticipate $\mathbf{m}_{t+1}$ at time $t+1$, where target $\mathbf{y}[\mathbf{u}_{t+1}] \in \mathbb{R}^{1\times |\mathcal{M}|}$.
\end{itemize}

\begin{figure*}[!h]
\centering
\includegraphics[width=0.95\linewidth, height=0.31\linewidth]{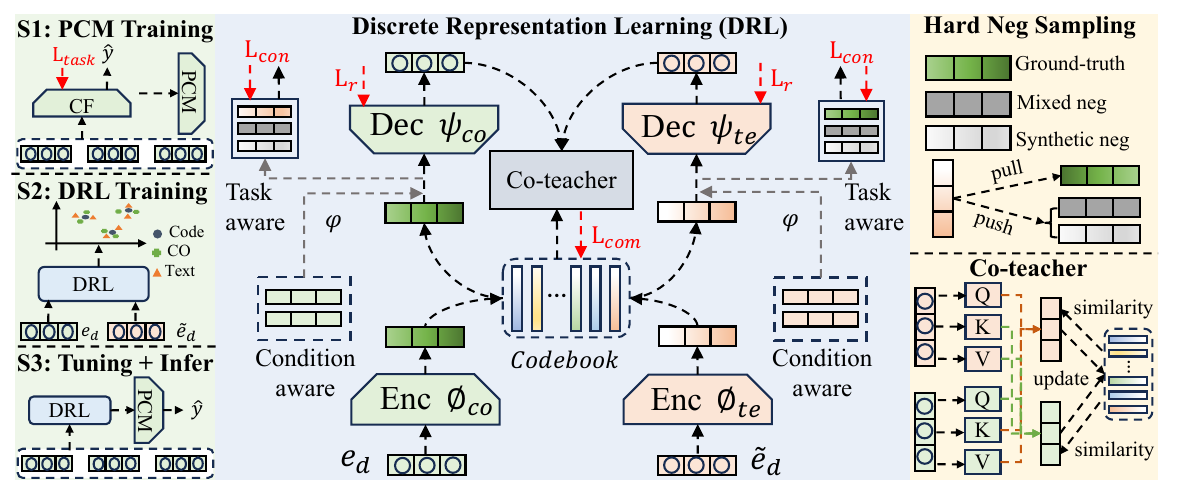}
\centering
\setlength{\abovecaptionskip}{-0.05cm}   
\setlength{\belowcaptionskip}{-0.1cm}   
\caption{Overview of \textit{UDC}. We pre-train the PCM to establish a robust CO space and then obtain CO and text representations for diseases using PCM and a selected PLM. Next, we train the DRL to align the text and CO signals, followed by fine-tuning the PCM for downstream tasks while keeping the DRL frozen. Q, K, and V denote the parameters for multi-head attention.} 
\label{fig:frame}
\vspace{-1em}
\end{figure*}

\noindent\textbf{Solution Overview.}
Our solution for enhancing healthcare prediction, particularly for rare diseases, unfolds through a structured three-step process. First, we develop a robust healthcare prediction model $\mathcal{F}_{\text{co}}(\cdot)$ by training on the entire dataset, which acts as the pre-trained collaborative model (PCM). However, this alone proves insufficient, as the resulting representations $\mathbf{E}_{\mathcal{D}}$ often fail to capture the nuances of rare diseases due to sparse co-occurrence. To address this, we choose a pre-trained language model (PLM), i.e. $\mathcal{F}_{\text{te}}(\cdot)$, and introduce a discrete representation learning (DRL) framework in the second stage, where we reconstruct these representations to ensure Text-CO signals alignment. Our key innovations lie in this phase, where we employ condition injection, contrastive learning, and co-teacher distillation to ensure that the discretized representations, incorporating both textual and collaborative signals, are distinct, task-aware, and aligned at the code level. Finally, in the fine-tuning \& inference stage, we freeze DRL to produce $\mathbf{\hat{E}}_\mathcal{D}$ that substitute the original embeddings $\mathbf{E}_\mathcal{D}$ and fine-tune $\mathcal{F}_{\text{co}}(\cdot)$, thereby significantly improving the model's capability to handle the challenging rare cases. The comprehensive framework is illustrated in Figure~\ref{fig:frame}.

\subsection{Discrete Disease Representation}
We employ discrete modeling to map disease representations onto discretized code vectors for reconstruction. Contrasted with  VAEs, VQ~\cite{lee2022autoregressive} process excels in compression and offers interpretability. 

\noindent\textbf{Pre-trained PCM \& PLM.}
Initially, we train a conventional healthcare prediction model optimized with commonly used {b}inary {c}ross-{e}ntropy (BCE)~\cite{jianggraphcare,dang2023uniform}, employing EHRs to construct collaborative representations for each medical entity. Formally, 
\begin{equation}\label{eq:1}
\setlength\abovedisplayskip{2pt}
\setlength\belowdisplayskip{2pt}
\mathbf{e}_{d} = \mathbf{E}_{\mathcal{D}}({d}), \quad
\mathbf{e}_{p} = \mathbf{E}_{\mathcal{P}}({p}), \quad
\mathbf{e}_{m} = \mathbf{E}_{\mathcal{M}}({m}), 
\end{equation}
\begin{equation}\label{eq:2}
\setlength\abovedisplayskip{2pt}
\setlength\belowdisplayskip{2pt}
 \mathcal{L}_{\text{task}} = \text{BCE}(\mathbf{y},\ \mathcal{F}_{\text{co}}(\mathbf{e}_{d}, \mathbf{e}_{p}, \mathbf{e}_{m}, \mathcal{T}_{k}; \theta)),
\end{equation}
where $\mathcal{F}_{\text{co}}(\cdot)$ can denote any PCM. Here, we opt for Transformer~\cite{vaswani2017attention} as the backbone. 
As evidenced in~\cite{zhao2023embedding,ali2023deep,wu2022survey}, $\mathcal{F}_{\text{co}}(\cdot)$ extracts interaction patterns, whereas embedding $\mathbf{E}$ encompasses rich CO similarities. 
Likewise, we choose a popular clinical pre-trained language model $\mathcal{F}_{\text{te}}(\cdot)$, i.e. Sap-BERT~\cite{liu2021self}, to serve as the PLM. Formally,
\begin{equation}\label{eq:3}
\setlength\abovedisplayskip{2pt}
\setlength\belowdisplayskip{2pt}
\mathrm{\Tilde{e}}_{d} = \mathbf{\Tilde{E}}_{\mathcal{D}}(\mathrm{T}({d})), \quad
\mathrm{\Tilde{e}}_{p} = \mathbf{\Tilde{E}}_{\mathcal{P}}(\mathrm{T}({p})), \quad
\mathrm{\Tilde{e}}_{m} = \mathbf{\Tilde{E}}_{\mathcal{M}}(\mathrm{T}({m})), 
\end{equation}
where $\mathbf{\Tilde{E}}$ signifies the embedding table of $\mathcal{F}_{\text{te}}(\cdot)$. 
We contrast the variations among various PCM and PLM backbones in Section~\ref{sec:plug}.

\noindent\textbf{Discrete Representation.}
Next, we consider mapping the disease encoding $\mathbf{e}_{d}$ and $\mathbf{\Tilde{e}}_{d}$ to a set of discrete codes using RQ-VAE~\cite{lee2022autoregressive}, a widely adopted VQ-VAE framework. In RQ-VAE, L-level codebooks are defined. For each code-level $l \in \{1,\cdots, L\}$, there exists a codebook $\mathcal{C}_{l}=\{\mathbf{c}_{i}\}^{|\mathcal{C}_{l}|}$. Subsequently, for disease $d$, the associated set of discrete codes is derived through the residual method. Formally, 
\begin{equation}\label{eq:4}
\left\{\begin{array}{l}
c_l=\underset{i}{\arg \min }\|\mathbf{r}_{l-1}-\mathbf{c}_i\|_2, \quad \mathbf{c}_i \in \mathcal{C}_{l}, \\
\mathbf{r}_l=\mathbf{r}_{l-1}-\mathbf{c}_{l},
\end{array}\right.
\end{equation}
where ${c}_{l}$ denotes the assigned code index from the $l$-th level codebook and $||\cdot||_{2}$ is 2-Norm. $\mathbf{r}_{l-1}$ is the semantic residual from the last level and we set $\mathbf{r}_{0}=\phi_{\text{co}}(\mathbf{e}_{d})$ or $\mathbf{\Tilde{r}}_{0}=\phi_{\text{te}}(\Tilde{\mathbf{e}}_{d})$, where $\phi$ is an MLP encoder layer. Finally, for each medical entity, we have the discrete PCM codes and discrete PLM codes, i.e., $\mathbf{e}_{d} \rightarrow \{\mathbf{c}_{1}, \mathbf{c}_{2}, \cdots, \mathbf{c}_{L}\}$, $\mathbf{\Tilde{e}}_{d} \rightarrow \{\mathbf{\Tilde{c}}_{1}, \mathbf{\Tilde{c}}_{2}, \cdots, \mathbf{\Tilde{c}}_{L}\}$. 
For efficiency, we utilize a shared codebook for both text and CO signals, i.e., $\mathbf{\Tilde{c}}_{l} \in \mathcal{C}_{l}$.
Then we get the encoded disease representation using the sum operation. Formally,
\begin{equation}\label{eq:5}
\setlength\abovedisplayskip{2pt}
\setlength\belowdisplayskip{2pt}
    \mathbf{z}_{d} = \sum_{l=1}^{L}\mathbf{c}_{l}, \quad \mathbf{\Tilde{z}}_{d} = \sum_{l=1}^{L}\mathbf{\Tilde{c}}_{l},
\end{equation}
where $\mathbf{z}_{d}$ and $\mathbf{\Tilde{z}}_{d}$ denote the discrete representation for a disease. In other words, we discretize the disease into the sum of various symptom codes, offering a more intuitive approach.


\subsection{Condition-aware Calibration}
Vanilla RQ-VAE typically proceeds to decode once the latent vector $\mathbf{z}_{d}$ is obtained. However, their efficacy in reconstructing samples with similar descriptions is limited. Mechanistically, vanilla RQ-VAE uses MSE loss to minimize overall reconstruction error, leading to identical "average" representations for similar text, sacrificing individual specificity~\cite{ji2023c2g2}.
This constraint significantly hampers their utility in healthcare scenes, where medical entities frequently share analogous descriptions yet possess distinct semantic nuances.
For instance, Type 1 and Type 2 diabetes may both be described as "diabetes without complications,…"(similar text) but they differ significantly in pathophysiology, warranting distinct representations in reconstruction. However, vanilla RQ-VAE produces similar representations for them due to overall MSE and similar text~\cite{ji2023c2g2,li2023survey}. 
To address this deficiency, we propose integrating external conditions, specifically diverse types of medical entities within the same visit, to modulate the quantization vector via normalization. This strategy aims to embed condition variations into the index map, thereby stimulating the decoder to produce a broader array of reconstructed representations. Formally,
\begin{equation}\label{eq:6}
\setlength\abovedisplayskip{2pt}
\setlength\belowdisplayskip{2pt}
\mathbf{f}_{d} = \text{MHA}_{\mathcal{P}}(\mathbf{e}_{p}^{d}, \mathbf{e}_{p}^{d}, 
\mathbf{e}_{p}^{d}) + \text{MHA}_{\mathcal{M}}(\mathbf{e}_{m}^{d}, \mathbf{e}_{m}^{d}, \mathbf{e}_{m}^{d}),
\end{equation}
where $\text{MHA}(\cdot)$ denotes the multi-head attention from Appendix~\ref{app:cond} and $\mathbf{f}_{d}$ is the condition representation. $\mathbf{e}_{p}^{d} \in \mathbf{E}_{\mathcal{P}}$ and $\mathbf{e}_{m}^{d} \in \mathbf{E}_{\mathcal{M}}$ denote the  entities for disease $d$ at the same visit. Then, we incorporate it in normalized form. Formally, for the CO branch,
\begin{equation}\label{eq:7}
\setlength\abovedisplayskip{2pt}
\setlength\belowdisplayskip{2pt}
\mathbf{z}_{d}=\varphi_\gamma(\mathbf{z}_d^{\text{old}}) \frac{\mathbf{f}_{d}-\mu(\mathbf{f}_{d})}{\sigma(\mathbf{f}_{d})}+\varphi_\beta(\mathbf{z}_d^{\text{old}}),
\end{equation}
where $\mathbf{z}_d^{\text{old}}$, as defined in Eq.~\ref{eq:5}, is labeled as "old" for clarity. $\mu$ and $\sigma$ denotes the mean and variation. $\varphi_\gamma$ and $\varphi_\beta$ signify the transformation matrix. This normalizing ensures that $\mathbf{f}$'s values fall within a similar range, which helps maintain consistency in the scale of the input features, thereby aiding in training stability and convergence without escalating the model's complexity. Likewise, we could obtain $\mathbf{\Tilde{z}}_{d}$ using $\mathbf{\Tilde{e}}_{p}^{d}$ and $\mathbf{\Tilde{e}}_{m}^{d}$.

\subsection{Task-aware Calibration}
While incorporating conditions can enhance the semantics of $\mathbf{z}_{d}$ for decoding, there remains a crucial gap: the model lacks awareness of downstream tasks. This awareness can help optimize model performance by guiding the learning process towards features that are most relevant to the healthcare task, leading to improved accuracy. In other words, we necessitate that the reconstructed representation not only mirrors the original one but also closely aligns with the target $\mathcal{S}_{d}$ in the subsequent visit ($\mathcal{S}_{d}\in \mathcal{D}$ for Diag Pred and $\mathcal{S}_{d} \in \mathcal{M}$ for Med Rec); otherwise, it remains distant. To achieve this objective, beyond conventional intra-domain (Text/CO signal) contrastive learning~\cite{CONVERT,li2024gslb}, we devise two distinct hard negative sampling to augment the contrastive training approach.  
Formally, using CO signal $\mathbf{z}_{d}$ as an example,
\begin{equation}\label{eq:8}
\setlength\abovedisplayskip{2pt}
\setlength\belowdisplayskip{2pt}
\mathcal{L}_{\text{intra}}=-\frac{1}{|\mathcal{D}|} \sum_{d=1}^{|\mathcal{D}|} \log [\frac{\exp (\mathbf{s}_{d} W \mathbf{z}_d)}{ \underbrace{\exp (\mathbf{s}_{d'} W \mathbf{z}_{d})}_{\text{synthetic}} + \sum_{j \neq d} \underbrace{\exp (\mathbf{s}_j W \mathbf{z}_d)}_{\text{intra-domain}}}],
\vspace{-1em}
\end{equation}
\begin{equation}\label{eq:9}
\setlength\abovedisplayskip{2pt}
\setlength\belowdisplayskip{2pt}
\mathcal{L}_{\text{inter}}=-\frac{1}{|\mathcal{D}|} \sum_{d=1}^{|\mathcal{D}|} \log [\frac{\exp (\mathbf{\Tilde{s}}_{d} W \mathbf{z}_d)}{ \underbrace{\exp (\mathbf{\Tilde{s}}_{d'} W \mathbf{z}_{d})}_{\text{synthetic}} + \sum_{j \neq d} \underbrace{\exp (\mathbf{\Tilde{s}}_j W \mathbf{z}_d)}_{\text{mixed-domain}}}],
\end{equation}
where $\mathbf{s}_{d}$ denotes $d$'s next-visit target representation, i.e., $\mathbf{s}_{d}=\sum_{d \in \mathcal{S}_{d}}\phi_{\text{co}}(\mathbf{e}_{d})$. $\mathbf{s}_{d'}$ denotes the synthetic disease representation acquired by randomly substituting the medical entities associated with the target $\mathcal{S}_{d}$. 
Likewise, we define $\mathbf{\Tilde{s}}_{d}=\sum_{d\in \mathcal{S}_{d}}\phi_{\text{te}}(\mathbf{\Tilde{e}}_{d})$.
Formally, we advance from both collaborative and textual standpoints,
\begin{equation}\label{eq:10}
\setlength\abovedisplayskip{2pt}
\setlength\belowdisplayskip{2pt}
\mathcal{L}_{\text{con}} = \mathcal{L}_{\text{intra}} + \mathcal{L}_{\text{inter}} + \mathcal{\Tilde{L}}_{\text{intra}} + \mathcal{\Tilde{L}}_{\text{inter}},
\end{equation}
where $\mathcal{\Tilde{L}}$ signifies the contrastive learning using $\mathbf{\Tilde{z}}_{d}$. This bidirectional learning ensures that the representations reconstructed by PCM and PLM not only encapsulate the relevance within the domain but also encompass the similarity of entities across domains.

\subsection{Co-teacher Distillation}
In the preceding sections, we transform both the CO and textual signals into discrete representations. However, this pipeline does not ensure semantic alignment between the two at the code level, leading to a domain gap that significantly impedes the subsequent Text$\rightarrow$CO mapping. 
To address this constraint, we introduce a co-teacher distillation that iteratively refines the same code by leveraging both text and CO signals. Specifically, for each code $\mathbf{c}_{i}$, we first retrieve the related diseases set $N_{i}^{l}$ and $\Tilde{N}_{i}^{l}$ in the collaborative and textual domain at the $l$-th level codebook. Subsequently, we combine their representations to obtain a holistic view $\mathbf{o}^{l}$. For clarity, we omit the superscript $l$. Formally, for $t$-th iteration,
\begin{equation}\label{eq:11}
\setlength\abovedisplayskip{2pt}
\setlength\belowdisplayskip{2pt}
\begin{aligned}
\mathbf{o}_i^{(t)}=\kappa \mathbf{o}_i^{(t-1)}+(1-\kappa)[\sum_{d \in N_i^{(t)}} &\frac{\mathbf{z}_{d}^{(t)}+\mathbf{\Tilde{b}}_{ d}^{(t)}}{2}+\sum_{d \in \Tilde{N}_i^{(t)}} \frac{\mathbf{\Tilde{z}}_{d}^{(t)}+\mathbf{b}_{d}^{(t)}}{2}], \\
\mathbf{b}^{(t)}_{d} =\text{MHA}(\mathbf{z}_{d}, \mathbf{\Tilde{z}}_{d}, \mathbf{\Tilde{z}}_{d}),& \quad  \mathbf{\Tilde{b}}^{(t)}_{d}= \text{MHA}(\mathbf{\Tilde{z}}_{d}, \mathbf{{z}}_{d}, \mathbf{{z}}_{d}), 
\end{aligned}
\end{equation}
where $\kappa$ refers to the decay rate and $\mathbf{b}$ extract the relationship between two views. Then, we employ an exponential moving average method to update $\mathbf{c}_{i}$. Formally,
\begin{equation}\label{eq:12}
\setlength\abovedisplayskip{2pt}
\setlength\belowdisplayskip{2pt}
\begin{aligned}
\mathbf{c}_i^{(t)}&=\mathbf{o}_i^{(t)} / \mathbf{n}_i^{(t)}, \\
\mathbf{n}_i^{(t)}=\kappa \mathbf{n}_i^{(t-1)}+(1-\kappa)&[\sum_{d \in N_i^{(t)}}\mathbf{z}_{d}^{(t)}+\sum_{d \in \Tilde{N}_i^{(t)}}\mathbf{\Tilde{z}}_{d}^{(t)}],
\end{aligned}
\end{equation}
where $\mathbf{n}_{i}$ are used for normalization. We also modify the commitment loss in RQ-VAE by utilizing the code vector $ \mathbf{\Tilde{z}}_{d}$ as a teacher to guide the encoder $\phi_{\text{co}}$. This modification aims for $\phi_{\text{co}}(\mathbf{e}_{d})$ to not only approximate $\mathbf{z}_{d}$ but also to converge towards $\mathbf{\Tilde{z}}_{d}$ at a ratio of 50\% in our setting, with $\alpha$ is the commitment weight. Formally, 
\begin{equation}\label{eq:13}
\setlength\abovedisplayskip{2pt}
\setlength\belowdisplayskip{2pt}
\begin{aligned}
\mathcal{L}_{\text {com}}= & \underbrace{\alpha\|\phi_{\text{co}}(\mathbf{e}_d)-\text{sg}[\mathbf{z}_d]\|_2^2}_{\text{origin}}+\underbrace{\frac{\alpha}{2}\|\phi_{\text{co}}(\mathbf{e}_d)-\text{sg}[\mathbf{\Tilde{z}}_d]\|_2^2}_{\text{new}} + \\
&\underbrace{\alpha\|\phi_{\text{te}}(\mathbf{\Tilde{e}}_d)-\text{sg}[\mathbf{\Tilde{z}}_i]\|_2^2}_{\text{origin}}+\underbrace{\frac{\alpha}{2}\|
\phi_{\text{te}}(\mathbf{\Tilde{e}}_d)-\text{sg}[\mathbf{{z}}_d]\|_2^2}_{\text{new}}
\end{aligned},
\end{equation}
where $\text{sg}$ denotes the stop gradient. 
This alignment compels the CO signal and the textual space to converge on the same symptom code at each discrete level and maintain the consistent code semantics, thereby facilitating subsequent representation substitution.

\vspace{-1em}
\subsection{Training \& Fine-tuning Strategy}
We outline the training objectives of the DRL and fine-tuning stages.

\noindent\textbf{Training Strategy.}
Our final optimization objective for DRL comprises reconstruction loss and the two preceding parts. Formally, 
\begin{equation}\label{eq:14}
\setlength\abovedisplayskip{2pt}
\setlength\belowdisplayskip{2pt}
\mathcal{L}_{\text{total}}=\underbrace{\|\mathbf{e}_d-\psi_{\text{co}}(\mathbf{z}_d)\|_2^2 + \|\mathbf{\Tilde{e}}_d-\psi_{\text{te}}(\mathbf{\Tilde{z}}_d)\|_{2}^{2}}_{\text {reconstruction loss} \ \mathcal{L}_\text{r}}+ \mathcal{L}_{\text{con}} + \mathcal{L}_{\text{com}},
\end{equation}
where $\psi$ denotes the MLP decoder for reconstruction. Once DRL is trained, it can be used as a mapping function to transform textual space into collaborative space. At this stage, we exclusively leverage data related to common diseases $\mathcal{D}_{\text{com}}$, as collaborative signals from rare diseases $\mathcal{D}_{\text{rar}}$ are considered unreliable. $\mathcal{D}_{\text{com}}$ and $\mathcal{D}_{\text{rar}}$ are splited according to Section~\ref{sec:expset}.

\noindent\textbf{Fine-tuning \& Inference.} 
Upon DRL alignment training completion, DRL can transform textual signals into collaborative signals. This enables us to utilize the textual description of rare diseases to supplant their original inferior collaborative signals. Formally, 
\begin{equation}\label{eq:15}
\setlength\abovedisplayskip{2pt}
\setlength\belowdisplayskip{2pt}
    \mathrm{\hat{e}}_{d} = 
    \begin{cases}        
    \psi_{\text{co}}[\varphi(\phi_{\text{te}}(\mathbf{\Tilde{e}}_{d}); \mathbf{e}_{p}^{d}, \mathbf{e}_{m}^{d})], & \text{if } d \in \mathcal{D}_{\text{rar}}\\
    \psi_{\text{co}}[\varphi(\phi_{\text{co}}(\mathbf{{e}}_{d}); \mathbf{e}_{p}^{d}, \mathbf{e}_{m}^{d})], & \text{if } d \in \mathcal{D}_{\text{com}}
    \end{cases}.
\end{equation}
Following this, we freeze DRL and $\mathbf{E}_\mathcal{D}$, and fine-tune $\mathcal{F}_{\text{co}}(\cdot)$ to capture updated  interaction patterns using Eq.~\ref{eq:2}. This step is crucial, as evidenced in Appendix~\ref{app:div}, since the prior $\mathcal{F}_{\text{co}}(\cdot)$ may not fully grasp interaction patterns with other medical entities owing to the data scarcity on rare diseases. For a new representation, it necessitates re-learning to enhance its effectiveness. Finally, we can integrate $\mathcal{F}_{\text{co}}(\cdot)$ and DRL for the estimation $\mathbf{\hat{y}}$. Formally,
\begin{equation}\label{eq:16}
\setlength\abovedisplayskip{2pt}
\setlength\belowdisplayskip{2pt}
\mathbf{\hat{y}} = \mathcal{F}_{\text{co}}(\mathbf{\hat{e}}_{d}, \mathbf{e}_{p}, \mathbf{e}_{m}, \mathcal{T}_{k}; \theta).
\end{equation}
Overall, through the three-step process, we can effectively map rare diseases onto the feature space of common diseases using textual knowledge as a bridge, thereby enhancing their semantic richness. A concise algorithm flow can be seen in Appendix~\ref{app:alg}.

\begin{table*}[!ht]\small
\centering
\setlength{\abovecaptionskip}{-0.05cm}   
\setlength{\belowcaptionskip}{-0.1cm}   
\caption{Performance comparison: Diagnosis Prediction. K=20.} %
\label{tab:diag}
\resizebox{0.75\textwidth}{!}{
\begin{tabular}{c|c|c|c|c||c|c|c|c||l|l|l|l} 
\toprule
Dataset       & \multicolumn{4}{c||}{MIMIC-III}    & \multicolumn{4}{c||}{MIMIC-IV}     & \multicolumn{4}{c}{eICU}            \\ 
\hline
Method        & Acc@K & Pres@K & AUPRC & AUROC & Acc@K & Pres@K & AUPRC & AUROC & Acc@K & Pres@K & AUPRC & AUROC  \\ 
\hline
Transformer~  & 0.2841      & 0.3144       & 0.2289    & 0.9174    & 0.3047     & 0.3420       & 0.2476   & 0.9591    & 0.6431         & 0.7716         & 0.6777      & 0.9667    \\
MICRON~        & 0.2735      & 0.3025       & 0.2130    & 0.9147    & 0.3081     & 0.3434       & 0.2098    & 0.9545    & 0.6308         & 0.7748         & 0.6781      & 0.9698        \\
Deepr~         & 0.2834      & 0.3132       & 0.2277    & 0.9113    & 0.2615     & 0.2904       & 0.1998    & 0.9396    & 0.6304         & 0.7620         & 0.6430      & 0.9584        \\
HITANet~  & 0.2917      & 0.3228       & 0.2309    & 0.9180    & 0.2996     & 0.3368       & 0.2432    & 0.9574    & 0.6517         & 0.7767         & 0.6773      & 0.9644    \\
RETAIN~        & 0.2920      & 0.3284       & 0.2509    & 0.9175    & 0.3078     & 0.3314       & 0.2337    & 0.9427    & 0.6576         & 0.7805         & 0.6879      & 0.9613         \\
GRAM~          & 0.3190      & 0.3559       & 0.2631    & 0.9182    & 0.3024     & 0.3513       & 0.2318    & 0.9591    & 0.6452         & 0.7891         & 0.6993      & 0.9711        \\
Dipole~       & 0.3183      & 0.3587       & 0.2631    & 0.9158    & 0.2968     & 0.3336       & 0.2395    & 0.9593    & 0.6585         & 0.7864         & 0.6677      & 0.9643        \\
StageNet~      & 0.3011      & 0.3375       & 0.2408    & 0.9188    & 0.3153     & 0.3440       & 0.2489    & 0.9593    & 0.6599         & 0.7936         & 0.6645      & 0.9702        \\
SHAPE~       & 0.3214      & 0.3531       & 0.2593    & 0.9226    & 0.3170     & 0.3540       & 0.2407    & 0.9564    & 0.6510         & 0.7779         & 0.6850      & 0.9676      \\
StratMed~       & 0.3076      & 0.3425       & 0.2434    & 0.9225    & 0.3137     & 0.3602      & 0.2595    & 0.9531    & 0.6449         & 0.7663         & 0.6710      & 0.9653        \\
MedPath~     & 0.3189      & 0.3490      & 0.2560    & 0.9224    & 0.3203     & 0.3616       & 0.2589    & 0.9620    & 0.6600         & 0.7947         & 0.7000      & 0.9714        \\
HAR~           & 0.3204      & 0.3532       & 0.2599    & 0.9193    & 0.3224     & 0.3642       & 0.2605    & 0.9628    & 0.6540         & 0.7910         & 0.6995      & 0.9720        \\
GraphCare~     & 0.3213      & 0.3529       & 0.2595    & 0.9203    & 0.3220     & 0.3635       & 0.2593    & 0.9620    & 0.6569         & 0.7788         & 0.6795      & 0.9694         \\
SeqCare~       & 0.3245      & 0.3547       & 0.2616    & 0.9213    & 0.3233     & 0.3668       & 0.2669    & 0.9632    & 0.6639         & 0.7996         & 0.7043      & 0.9727        \\
RAREMed~       & 0.3208      & 0.3521       & 0.2596    & 0.9192    & 0.3153     & 0.3527       & 0.2390    & 0.9557   & 0.6572         & 0.7792         & 0.6768      & 0.9692       \\
\hline
\textit{UDC}  & \textbf{0.3377}      & \textbf{0.3713}       & \textbf{0.2737}    & \textbf{0.9256}    & \textbf{0.3324}     & \textbf{0.3707}       & \textbf{0.2735}    & \textbf{0.9657}    & \textbf{0.6724}         & \textbf{0.8070}         & \textbf{0.7140}      & \textbf{0.9736} \\
\bottomrule
\end{tabular}}
\vspace{-1em}
\end{table*}
\begin{table*}[!ht]\small
\centering
\setlength{\abovecaptionskip}{-0.05cm}   
\setlength{\belowcaptionskip}{-0.1cm}   
\caption{Performance comparison: Medication Recommendation.} 
\label{tab:medp}
\resizebox{0.75\textwidth}{!}{
\begin{tabular}{c|c|c|c|c||c|c|c|c||l|l|l|l} 
\toprule
Dataset       & \multicolumn{4}{c||}{MIMIC-III}    & \multicolumn{4}{c||}{MIMIC-IV}     & \multicolumn{4}{c}{eICU}            \\ 
\hline
Method        & Jaccard & F1-score & AUPRC & AUROC & Jaccard & F1-score & AUPRC & AUROC & Jaccard & F1-score & AUPRC & AUROC  \\ 
\hline
Transformer~  & 0.5012      & 0.6556       & 0.7671    & 0.9440    & 0.4635     & 0.6203       & 0.7305    & 0.9402            & 0.1159         & 0.3504      & 0.3138 & 0.9147    \\
MICRON~        & 0.4937      & 0.6501       & 0.7651    & 0.9307    & 0.4608     & 0.6123       & 0.7283    & 0.9362           & 0.0703         & 0.2349      & 0.2561    & 0.9017      \\
SafeDrug~      & 0.4859      & 0.6403       & 0.7367    & 0.9331    & 0.4569     & 0.6086       & 0.7293    & 0.9378            & 0.1061         & 0.4274      & 0.3036    & 0.9181      \\
RETAIN~        & 0.5049      & 0.6601       & 0.7680    & 0.9448    & 0.4646     & 0.6174       & 0.7364    & 0.9414    & 0.1181         & 0.4736         & 0.2835      & 0.9064         \\
GRAM~          & 0.4994      & 0.6537       & 0.7607    & 0.9435    & 0.4624     & 0.6155       & 0.7385    & 0.9424    & 0.0983         & 0.3166         & 0.2908      & 0.9168        \\
GAMENet~       & 0.5074      & 0.6612       & 0.7724    & 0.9456    &0.4655     & 0.6181       & 0.7399    & 0.9425            & 0.1093         & 0.4165      & 0.2936    & 0.9103    \\
COGNet~         & 0.5114      & 0.6614       & 0.7774    & 0.9470    & 0.4612     & 0.6125       & 0.7271    & 0.9356    & 0.1166         & 0.3528        & 0.3237      & 0.9147        \\
StageNet~      & 0.5013      & 0.6494       & 0.7519    & 0.9358    & 0.4679     & 0.6201       & 0.7404    & 0.9424    & 0.1337         & 0.2303         & 0.3075      & 0.9201        \\
VITA~       & 0.5146      & 0.6671       & 0.7781    & 0.9469    & 0.4715     & 0.6219       & 0.7486    & 0.9424     & 0.1218         & 0.3640         & 0.3223      & 0.9157      \\
MoleRec~       & 0.5080      & 0.6624       & 0.7719    & 0.9451    & 0.4720     & 0.6254       & 0.7473    & 0.9411    & 0.1123         & 0.3609         & 0.3280      & 0.9219     \\
DEPOT~       & 0.5135      & 0.6697       & 0.7745    & 0.9466   & 0.4780     & 0.6298       & 0.7534    & 0.9465    & 0.1367         & 0.3875         & 0.3276      & 0.9134        \\
SHAPE~      & 0.5155      & 0.6678       & 0.7788    & 0.9469  & 0.4830     & 0.6347       & 0.7486    & 0.9475    & 0.1338         & 0.4056         & 0.3123      & 0.9154        \\
StratMed~       & 0.5070      & 0.6612       & 0.7724    & 0.9456    & 0.4719     & 0.6249       & 0.7446    & 0.9446    & 0.1223         & 0.3791         & 0.3031      & 0.9138        \\
HAR~           &0.5126  & 0.6652 & 0.7758 & 0.9465    & 0.4805     & 0.6311       & 0.7539    & 0.9475    & 0.1257         & 0.4595         & 0.3153      & 0.9140        \\
GraphCare~      & 0.5167      & 0.6700       & 0.7805    & 0.9471   & 0.4816     & 0.6363       & 0.7576    & \textbf{0.9486}    & 0.1252         & 0.4534         & 0.3107      & 0.9162         \\
RAREMed~       & 0.5134      & 0.6653       & 0.7786    & 0.9453    & 0.4794     & 0.6317       & 0.7496    & 0.9443     & 0.1304         & 0.4315         & 0.3119      & 0.9156        \\
\hline
\textit{UDC}  & \textbf{0.5261}      & \textbf{0.6761}       & \textbf{0.7833}    & \textbf{0.9483}    & \textbf{0.4912}     & \textbf{0.6404}       & \textbf{0.7580}    & \textbf{0.9486}    & \textbf{0.1443}         & \textbf{0.4986}         & \textbf{0.3296}      & \textbf{0.9227} \\
\bottomrule
\end{tabular}}
\vspace{-1em}
\end{table*}
\section{Experiments}\label{sec:exp}
We first outline the necessary setup and then present the analysis.
\vspace{-1em}
\subsection{Experimental Setup}\label{sec:expset}
\noindent\textbf{Datasets \& Baselines.}
Our experiments are conducted on three popular healthcare datasets: MIMIC-III~\cite{johnson2016mimic}, MIMIC-IV~\cite{johnson2020mimic}, and eICU~\cite{pollard2018eicu}. Detailed statistics for these datasets are summarized in Appendix~\ref{app:sta}.
Textual knowledge is extracted by parsing EHR entities according to the internationally recognized ICD and ATC  systems~\cite{garcia2023association} to obtain corresponding textual descriptions. We retain patients with more than one visit in MIMIC-III and eICU, while for MIMIC-IV, we include patients with two or more visits.

We select advanced baselines for comparison. Specifically, for both tasks, we include Transformer~\cite{vaswani2017attention}, MICRON~\cite{wu2022conditional}, RETAIN~\cite{choi2016RETAIN}, GRAM~\cite{choi2017gram}, StageNet~\cite{gao2020stagenet}, SHAPE~\cite{liu2023SHAPE}, StratMed~\cite{li2024stratmed}, 
HAR~\cite{wang2023stage}, GraphCare~\cite{jianggraphcare}, and RAREMed~\cite{zhao2024leave}. For Diag Pred, we further incorporate HITANet~\cite{luo2020HITANet}, Deepr~\cite{nguyen2016mathtt}, Dipole~\cite{ma2017dipole}, 
MedPath~\cite{ye2021medpath}, SeqCare~\cite{xu2023seqcare} as specialized baselines. In Med Rec, additional baselines such as SafeDrug~\cite{yang2021safedrug}, GAMENet~\cite{shang2019gamenet}, 
COGNet~\cite{yang2021change}, VITA~\cite{kim2024vita}, MoleRec~\cite{yang2023molerec}, and 
DEPOT~\cite{zhao2024enhancing}, are included, given their distinctive designs and strong performance. Transformer, RETAIN, HITANet, Deepr, StageNet, RAREMed, and SHAPE are sequence-based approaches, while GRAM, GAMENet, MoleRec, MICRON, DEPOT, StratMed, COGNet, and VITA further integrate EHR graphs to enhance representation. MedPath, HAR, SeqCare, and GraphCare leverage external knowledge to improve performance. RAREMed and SeqCare incorporate tailored reconstruction tasks and denoising techniques specifically designed for rare diseases.

\noindent\textbf{Implementation Details \& Evaluations.}
To ensure fairness, following~\cite{jianggraphcare,zhao2024enhancing}, all algorithms use an embedding dimension of 128. We employ the AdamW optimizer with a learning rate of 1e-3 for Diag Pred and 2e-4 for Med Rec. The batch size is set to 16. The epochs for the DRL and fine-tuning stages are set at 50  and 50, respectively.
Following RQ-VAE, we configure the code layer $L=4$, meaning each disease is represented by four codes. The codebook size $|\mathcal{C}_{l}|$ and commitment weight $\alpha$, which are crucial hyperparameters, are set to 64 and 0.25, respectively. Their effects are evaluated in Appendix~\ref{sec:4.4}.
Following the Pareto principle and previous research~\cite{zhang2023deep}, we classify diseases appearing in $20\%$ or more cases as common $\mathcal{D}_{\text{com}}$, with all others considered rare $\mathcal{D}_\text{{rar}}$. The impact of varying  thresholds $\eta$ is further explored in Appendix~\ref{sec:4.4}.

For data partitioning, we follow established practices~\cite{zhao2024enhancing,wang2023stage,xu2023seqcare} by dividing the datasets into training, validation, and test sets in a 6:2:2 ratio. For Diag Pred, we use Acc@K, Pres@K, AUPRC, and AUROC for evaluation. Here K=20, different values are discussed in Section~\ref{sec:4.3.2}. For Med Rec, we assess using Jaccard, F1-score, PRAUC, and AUROC. These metrics are selected for their significant clinical relevance and comprehensive assessment~\cite{ali2023deep,jianggraphcare}. 
\vspace{-1em}

\subsection{Overall Performance}
As depicted in Tables~\ref{tab:diag}-\ref{tab:medp}, our proposed UDC achieves the best performance across all scenarios, despite only utilizing the relatively weak Transformer as the PCM.
Regarding the baselines, we observe that the sequence-based methods, such as SHAPE and DEPOT significantly outperform GRAM, underscoring the importance of capturing temporal patterns. COGNet and VITA are Transformer variants that leverage medical priors, like EHR graphs, resulting in notable enhancements over pure Transformer.
GraphCare, MedPath, and SeqCare distinguish themselves by leveraging external knowledge graphs to enrich the inherent entity semantics. Nevertheless, the absence of adequate denoising measures hinders their effectiveness.
While RAREMed introduces pre-trained tasks to address the cold-start issue, its overall predictive capacity remains relatively modest. Observations suggest a potential decline for common disease prediction, as detailed in Section~\ref{sec:4.3.3}.

Concerning the tasks, Diag Pred is more challenging than Med Rec, as the former requires recalling and ranking a broader range of medical entities. UDC, GraphCare, and SeqCare demonstrate greater robustness, as they not only rely on CO signals but also leverage semantic associations between items from the external knowledge. The broader Diag Pred benefits more from the external knowledge effects in the sampling process, leading to a 3\% Acc@K improvement in MIMIC-IV. Our observations indicate that eICU demonstrates enhanced performance in Diag Pred, likely due to the smaller disease size, which results in greater similarity among diseases across consecutive periods. MICRON's performance on MIMIC-III and eICU is constrained in both tasks due to its requirement for at least two visit lengths, which limits the available data.
StratMed does not reproduce its success from Med Rec on Diag Pred. This disparity could stem from the drug interaction graph it introduced not being suitable for the Diag Pred.

Considering the datasets, MIMIC-IV is the most challenging, as it exhibits more complex  entity interactions, reflected in the larger data volume and higher sparsity. Additionally, the MIMIC-IV data presents a more imbalanced distribution, as shown in Figure~\ref{fig:motiv}. Most algorithms, such as StratMed, Dipole, and DEPOT, experience noticeable performance degradation on this dataset. 
Despite incorporating external knowledge, as seen in GraphCare and HAR, their approaches overlook the domain gap between this knowledge and the CO signal, potentially leading to negative transfer. Meanwhile, the lack of standard EHR coding in the eICU dataset leads to significant gaps in external knowledge, diminishing the advantages of these baselines. 
Conversely, UDC directly leverages the text of eICU records and aligns CO signals with textual knowledge without requiring additional indexing, effectively alleviating this issue.


\subsection{Model Analysis and Robust Testings}
Without loss of generalization, we conduct various robustness experiments on MIMIC-III to validate our efficacy.

\begin{table}[!h]\small
\centering
\setlength{\abovecaptionskip}{-0.05cm}   
\setlength{\belowcaptionskip}{-0.1cm}   
\caption{Ablation study. UDC-NCO does not incorporate condition-aware calibration. UDC-NT removes task-aware calibration. UDC-NM only leverages synthetic negative sampling. UDC-NS only utilizes mixed-domain negative sampling.  UDC-NCD performs updates similar to RQ-VAE without using co-teacher distillation.
} 
\label{tab:aba}
\resizebox{0.48\textwidth}{!}{
\begin{tabular}{c|c|ccccc||c} 
\hline
Algorithms                & Metric & -NCO  & -NT  & -NM  & -NS & -NCD & UDC  \\ 
\hline
\multirow{2}{*}{Diag Pred}      & Acc@K  & 0.3276 & 0.3297 & 0.3301 & 0.3318     & 0.3288   & \textbf{0.3377}       \\
                          & Pres@K    & 0.3606 & 0.3600 & 0.3608 & 0.3620    & 0.3591   & \textbf{0.3713}        \\ 
\hline\hline
\multirow{2}{*}{Med Rec} & Jaccard & 0.5176 & 0.5179  & 0.5205 & 0.5183    & 0.5171   & \textbf{0.5261}      \\
                          & F1-score  & 0.6703 & 0.6705  & 0.6709 & 0.6706    & 0.6689   & \textbf{0.6761}      \\
                          
\hline
\end{tabular}}
\vspace{-1em}
\end{table}
\subsubsection{Ablation Study}
We conduct ablation experiments to validate the efficacy of sub-modules. As shown in Table~\ref{tab:aba}, UDC-CO, which lacks the condition-aware modeling between the disease and visit components, is the limited-effective configuration, with a substantial 3\% drop in Diag Pred. This absence causes disease, akin to textual descriptions, to be challenging for the model to differentiate, thereby leading to a blurred decision boundary.
While UDC-NT has little impact on the reconstruction ability, it fails to impose effective constraints on the representation space. 
Directly applying this representation to downstream tasks proves challenging, necessitating additional training during the fine-tuning phase, yet achieving equivalent performance remains elusive.
When contrasted with UDC-NT, both UDC-NM and UDC-NS exhibit enhanced performance, attributed to their capability to enhance the model's individual discernment by integrating hard negative instances.
UDC-NCD, akin to RQ-VAE in codebook update, experiences a 2\% degradation due to domain gaps between text and CO spaces. This disparity could result in a significant negative transfer.
Overall, the results validate the essential contributions of the key sub-modules.
\begin{figure}[!h] 
\vspace{-1em}
\centering
\subfigure[Diag Pred (Acc@K)]{
\begin{minipage}[t]{0.48\linewidth}
\centering
\includegraphics[width=\linewidth,height=0.7\linewidth]{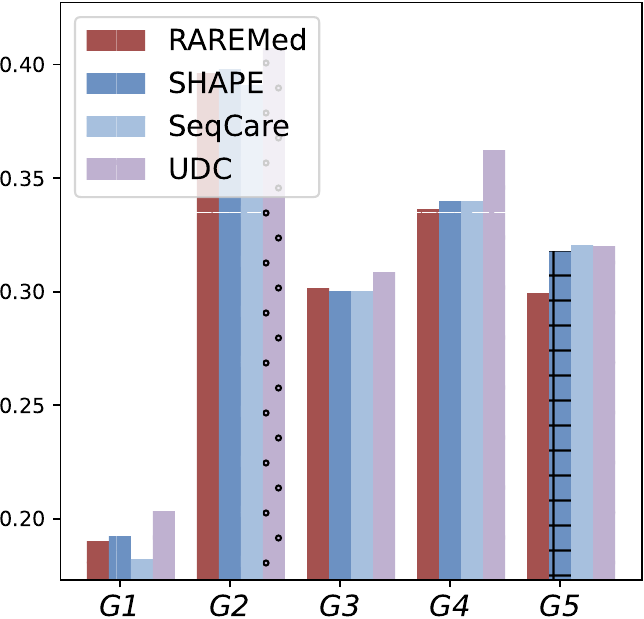}
\label{fig:gr:diag:acc}
\end{minipage}%
}%
\subfigure[Med Rec (Jaccard)]{
\begin{minipage}[t]{0.48\linewidth}
\centering
\includegraphics[width=\linewidth,height=0.7\linewidth]{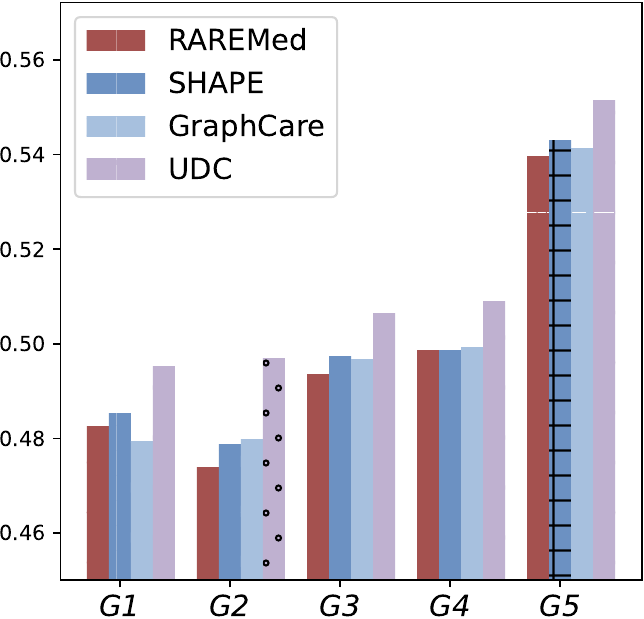}
\label{fig:gr:med:jac}
\end{minipage}%
}%
\centering
\setlength{\abovecaptionskip}{-0.05cm}   
\setlength{\belowcaptionskip}{-0.1cm}   
\caption{Group Analysis.}
\label{fig:gr}
\vspace{-1em}
\end{figure} 

\subsubsection{Group Analysis}\label{sec:4.3.3}
To examine the model's performance on rare diseases, we conduct a group-level analysis.
Specifically, in Diag Pred, diseases are categorized into five prevalence groups: 0-20\% (G1), 20-40\%(G2), 40-60\%(G3), 60-80\%(G4), and 80-100\%(G5), where G1 is the rarest disease group.
As shown in Figure~\ref{fig:gr:diag:acc}, the model's efficacy in Diag Pred generally exhibits a positive correlation with the sparsity of the disease groups, with G2-G5 significantly outperforming G1.  However, the performance of the G5 is not optimal, likely due to the low clinical significance of high-frequency diseases in Diag Pred; for instance, fever can indicate multiple underlying health risks.
While RAREMed surpasses other baselines in G1 and G3, it compromises accuracy for common diseases.  UDC  exhibits the most notable boost in G1-G4, showcasing that our innovations excel at enhancing performance for rare diseases.

For the Med Rec, we further analyze the predictive performance for patient groups with various rare diseases. 
More precisely, we identify the rarest disease for each patient and allocate them to the corresponding group based on that rarity.
Figure~\ref{fig:gr:med:jac} indicates that recommendation performance for G1-G3 is limited, as fewer medications co-occur with their disease entities, leading to weaker disease-medication CO signals. Both SHAPE and RAREMed suffer from this issue.
While GraphCare attempts to mitigate this problem by leveraging external knowledge, it fails to fully bridge the domain gap during the knowledge fusion and suffers from the potential knowledge noise. In contrast, UDC explicitly optimizes code-level alignment in DRL, facilitating bidirectional alignment of CO signals and textual knowledge, which leads to remarkable improvements.

In general, group-level analyses confirm that UDC significantly outperforms other baselines in managing rare diseases, essential for effective clinical decision support.

\vspace{-0.2em}
\begin{figure*}[!ht] 
\centering
\subfigure[Diag Pred (Acc@K)]{
\begin{minipage}[t]{0.21\linewidth}
\centering
\includegraphics[width=\linewidth,height=0.7\linewidth]{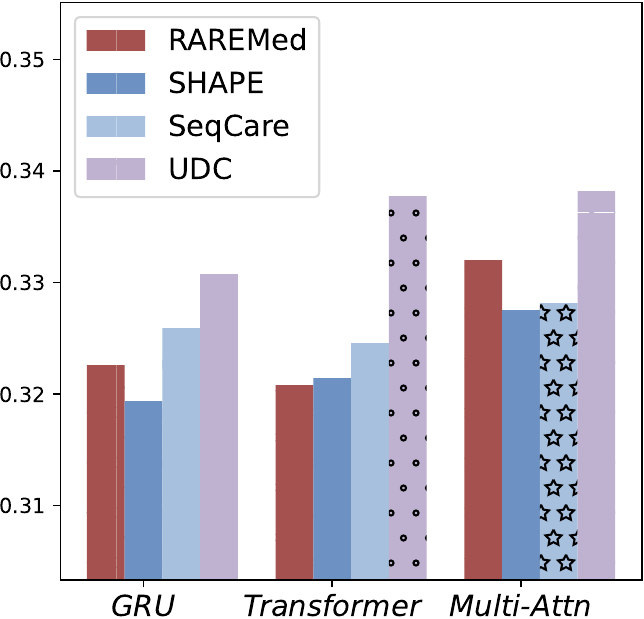}
\label{fig:plug-in:pcm:dia:acc}
\end{minipage}%
}%
\subfigure[Diag Pred (Pres@K)]{
\begin{minipage}[t]{0.21\linewidth}
\centering
\includegraphics[width=\linewidth,height=0.7\linewidth]{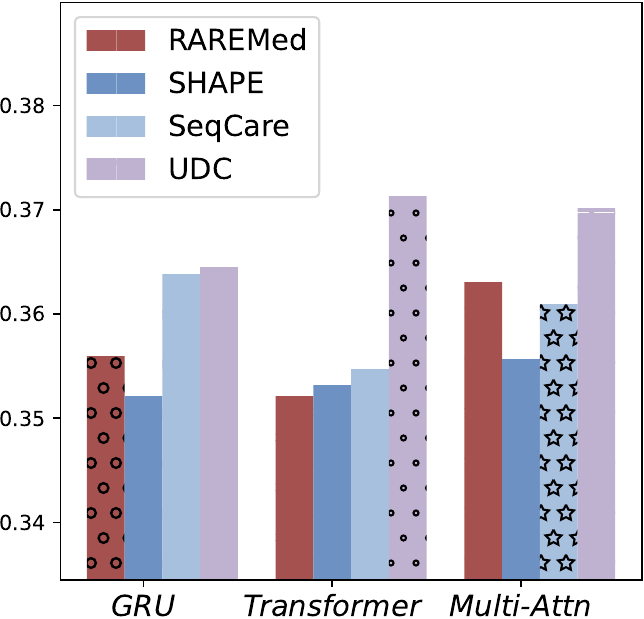}
\label{fig:plug-in:pcm:dia:pre}
\end{minipage}%
}%
\subfigure[Med Rec (Jaccard)]{
\begin{minipage}[t]{0.21\linewidth}
\centering
\includegraphics[width=\linewidth,height=0.7\linewidth]{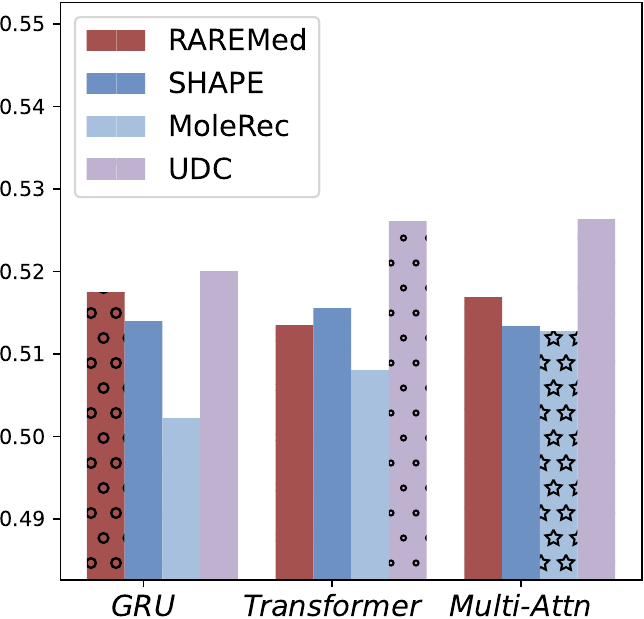}
\label{fig:plug-in:pcm:med:jac}
\end{minipage}%
}%
\subfigure[Med Rec (F1-score)]{
\begin{minipage}[t]{0.21\linewidth}
\centering
\includegraphics[width=\linewidth,height=0.7\linewidth]{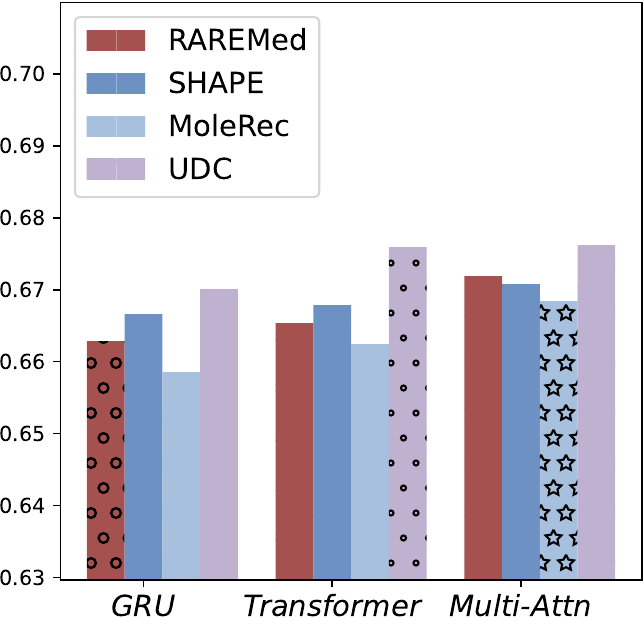}
\label{fig:plug-in:pcm:med:f1}
\end{minipage}%
}%
\centering
\setlength{\abovecaptionskip}{-0.15cm}   
\setlength{\belowcaptionskip}{-0.1cm}   
\caption{Plug-in Application (Diverse PCM). We choose MoleRec, SHAPE, RAREMed, and SeqCare, as they are flexible to PCM.}
\label{fig:plug-in:pcm}
\vspace{-1em}
\end{figure*} 
\begin{figure*}[!ht] 
\centering
\subfigure[Diag Pred (Acc@K)]{
\begin{minipage}[t]{0.21\linewidth}
\centering
\includegraphics[width=\linewidth,height=0.7\linewidth]{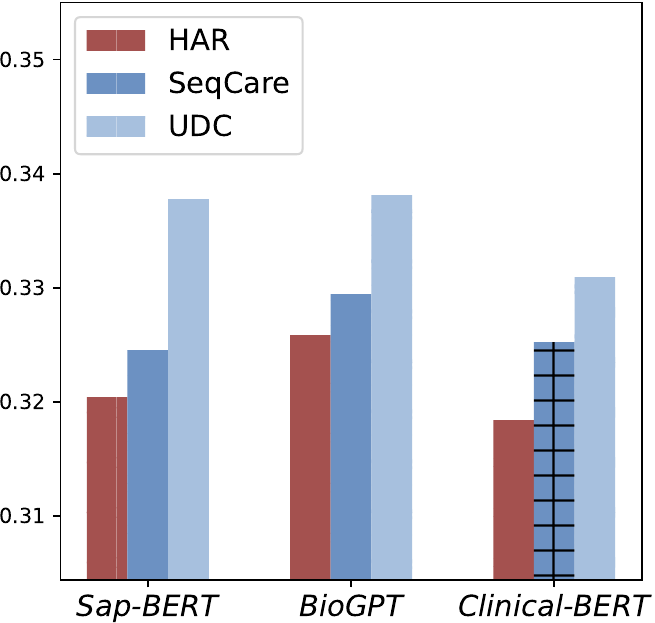}
\label{fig:plug-in:plm:dia:acc}
\end{minipage}%
}%
\subfigure[Diag Pred (Pres@K)]{
\begin{minipage}[t]{0.21\linewidth}
\centering
\includegraphics[width=\linewidth,height=0.7\linewidth]{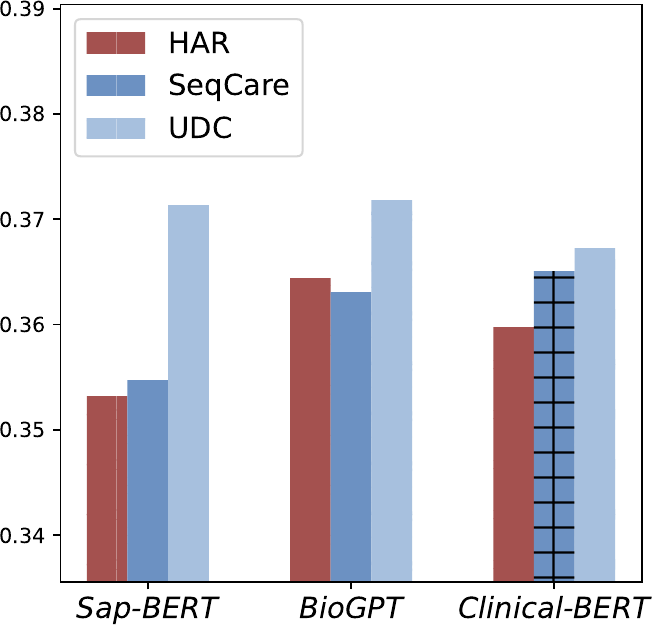}
\label{fig:plug-in:plm:dia:pre}
\end{minipage}%
}%
\subfigure[Med Rec (Jaccard)]{
\begin{minipage}[t]{0.21\linewidth}
\centering
\includegraphics[width=\linewidth,height=0.7\linewidth]{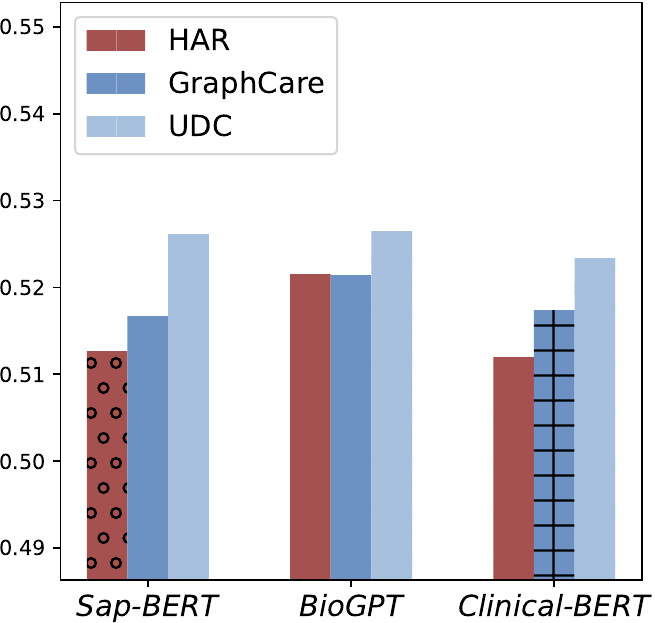}
\label{fig:plug-in:plm:med:jac}
\end{minipage}%
}%
\subfigure[Med Rec (F1-score)]{
\begin{minipage}[t]{0.21\linewidth}
\centering
\includegraphics[width=\linewidth,height=0.7\linewidth]{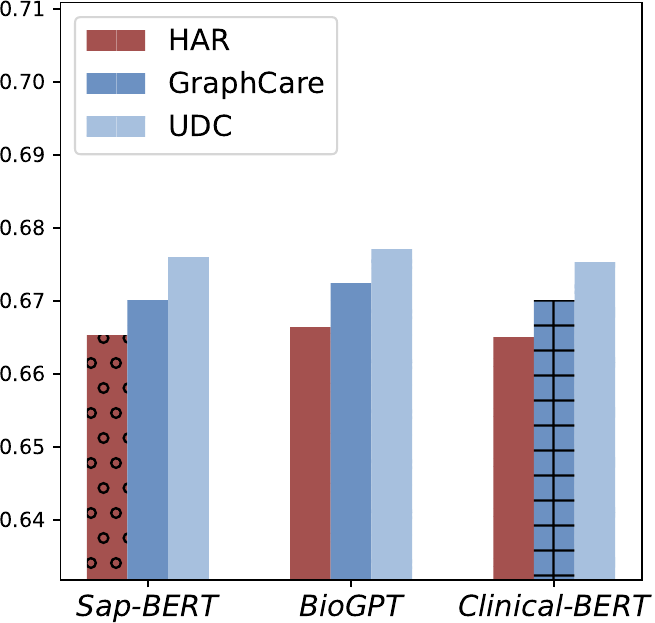}
\label{fig:plug-in:plm:med:f1}
\end{minipage}%
}%
\centering
\setlength{\abovecaptionskip}{-0.15cm}   
\setlength{\belowcaptionskip}{-0.1cm}   
\caption{Plug-in Application (Diverse PLM). We select  HAR, GraphCare, and SeqCare that utilize external knowledge.}
\label{fig:plug-in:plm}
\vspace{-1em}
\end{figure*} 
\subsubsection{Plug-in Application}\label{sec:plug}
We examine the extensibility of UDC.

\noindent\textbf{Diverse PCM.}
For the PCM, we select three modern methods—GRU, Transformer, and Multi-head Attention—due to their widespread use in sequence-based healthcare baselines~\cite{shang2019gamenet,ma2017dipole,yang2023molerec}. As shown in Figure~\ref{fig:plug-in:pcm}, RAREMed has larger fluctuations, likely due to its explorations of three CO signals, maximizing its advantage from PCM. UDC demonstrates robust performance with various sophisticated PCM. The improvement in Multi-head Attention variants results from their significant CO advancements and convergence toward a more precise subspace during DRL alignment. This superior convergence contributes to an overall boost in model performance.

\noindent\textbf{Diverse PLM.}
Similarly, for the PLM, we evaluate the integration of both BioGPT~\cite{luo2022biogpt} and Clinical-BERT~\cite{wang2023optimized}. Understanding the textual semantics encoded in clinical notes is another crucial aspect of the DRL, as it can capture similarities between entities that may not be evident from the EHRs alone. Compared to the Sap-BERT and Clinical-BERT, the BioGPT, which is fine-tuned on larger medical-domain corpora, possesses more semantic representations. Furthermore, the larger parameter capacity of BioGPT enables it to obtain an even more robust alignment of the DRL module, leading to notable performance gains when integrated into UDC.

\begin{figure}[!h] 
\centering
\subfigure[Before DRL (Diag Pred)]{
\begin{minipage}[t]{0.48\linewidth}
\centering
\includegraphics[width=0.9\linewidth,height=0.65\linewidth]{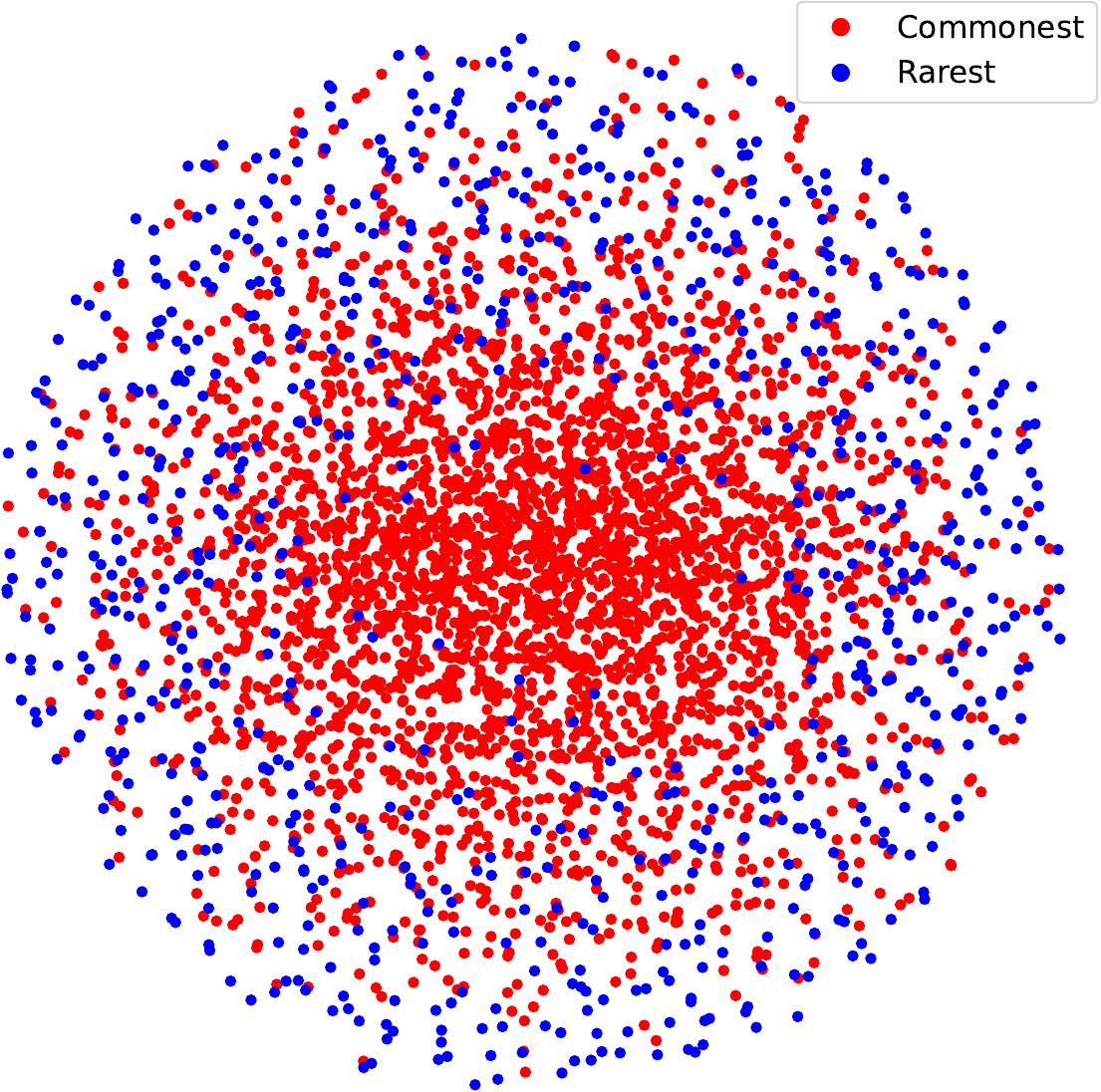}
\label{fig:case:diag:before}
\end{minipage}%
}%
\subfigure[After DRL (Diag Pred)]{
\begin{minipage}[t]{0.48\linewidth}
\centering
\includegraphics[width=0.9\linewidth,height=0.65\linewidth]{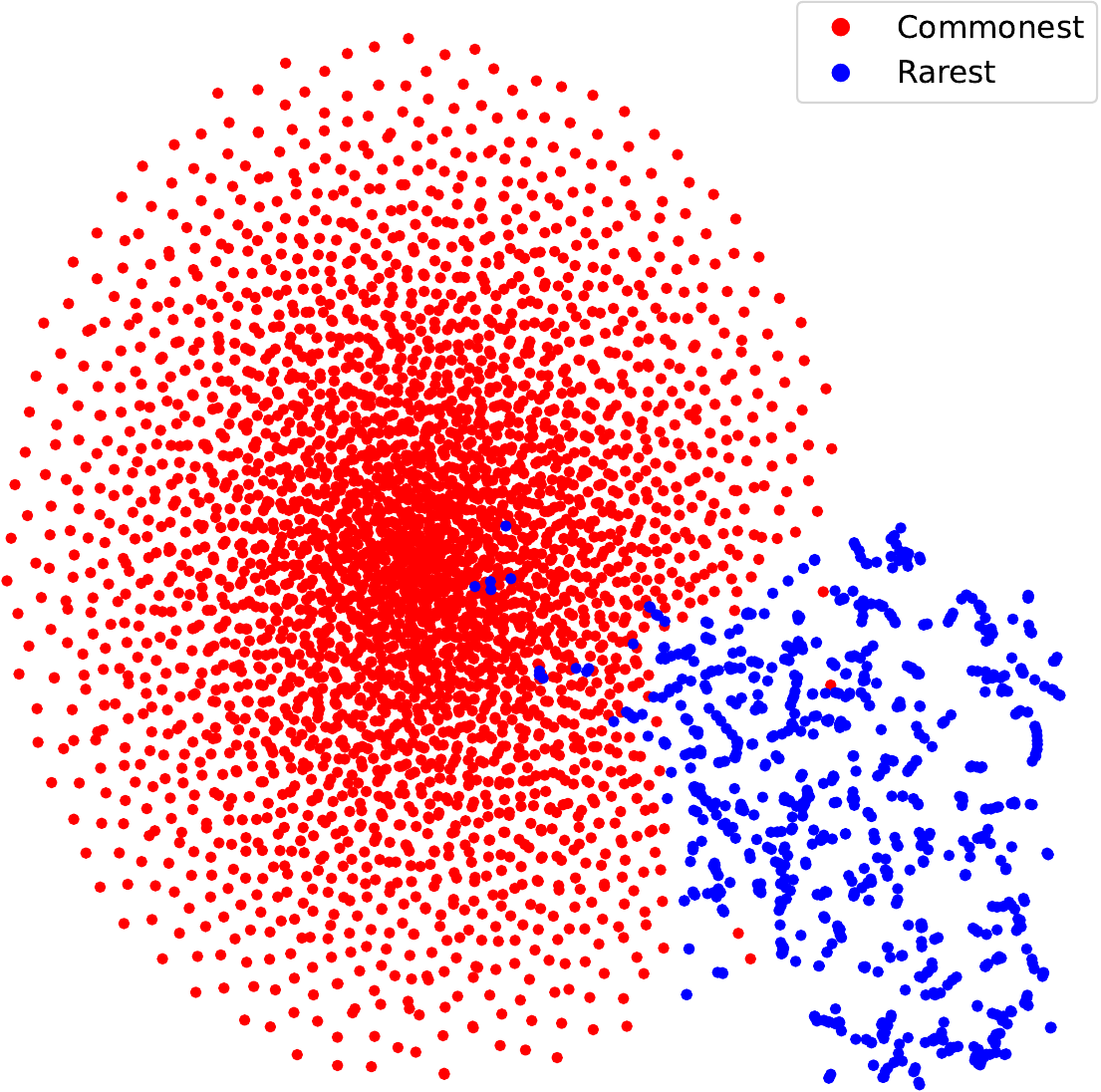}
\label{fig:case:diag:after}
\end{minipage}%
}%
\vfill
\subfigure[Before DRL (Med Rec)]{
\begin{minipage}[t]{0.48\linewidth}
\centering
\includegraphics[width=0.9\linewidth,height=0.65\linewidth]{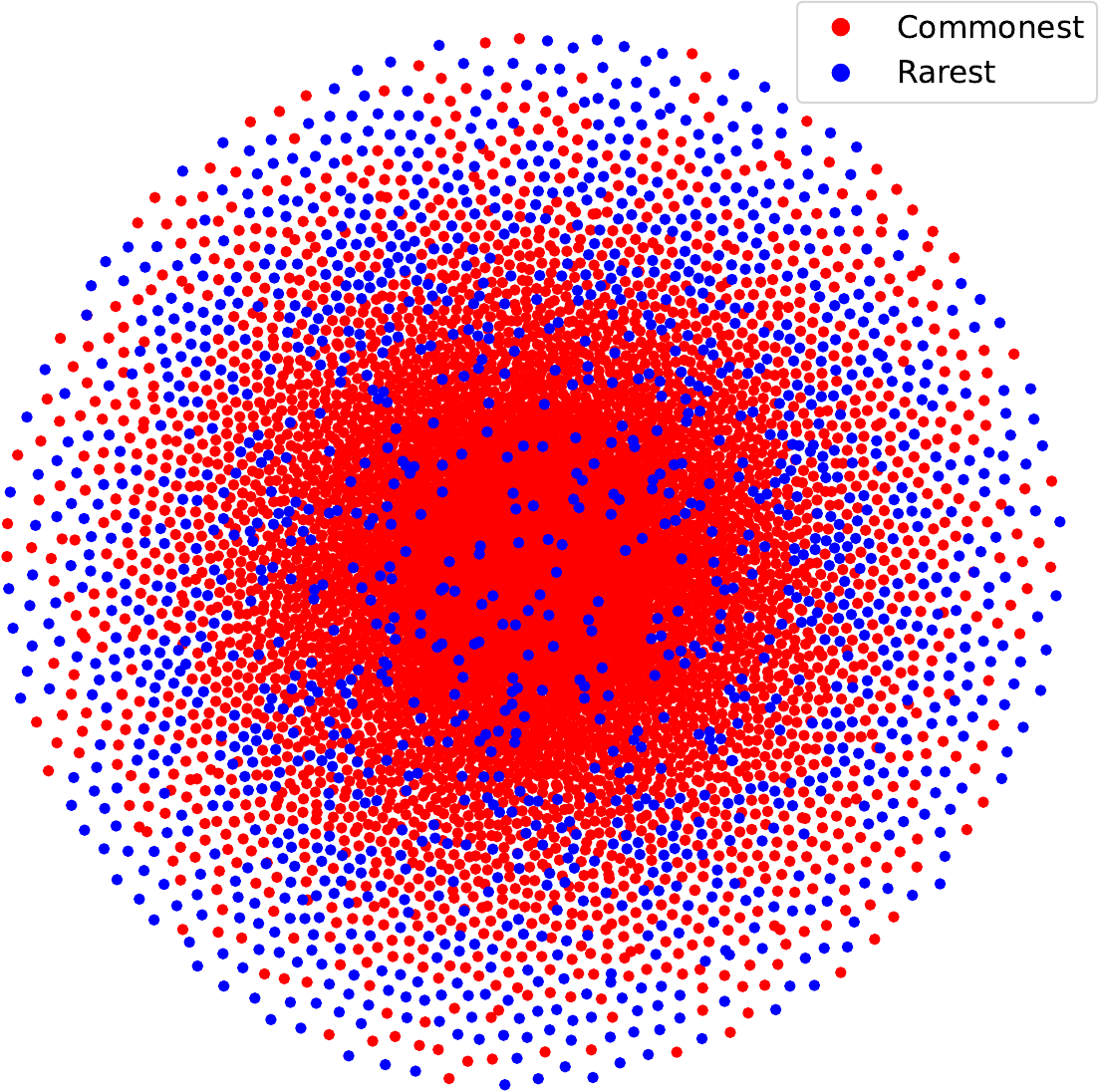}
\label{fig:case:med:before}
\end{minipage}%
}%
\subfigure[After DRL (Med Rec)]{
\begin{minipage}[t]{0.48\linewidth}
\centering
\includegraphics[width=0.9\linewidth,height=0.65\linewidth]{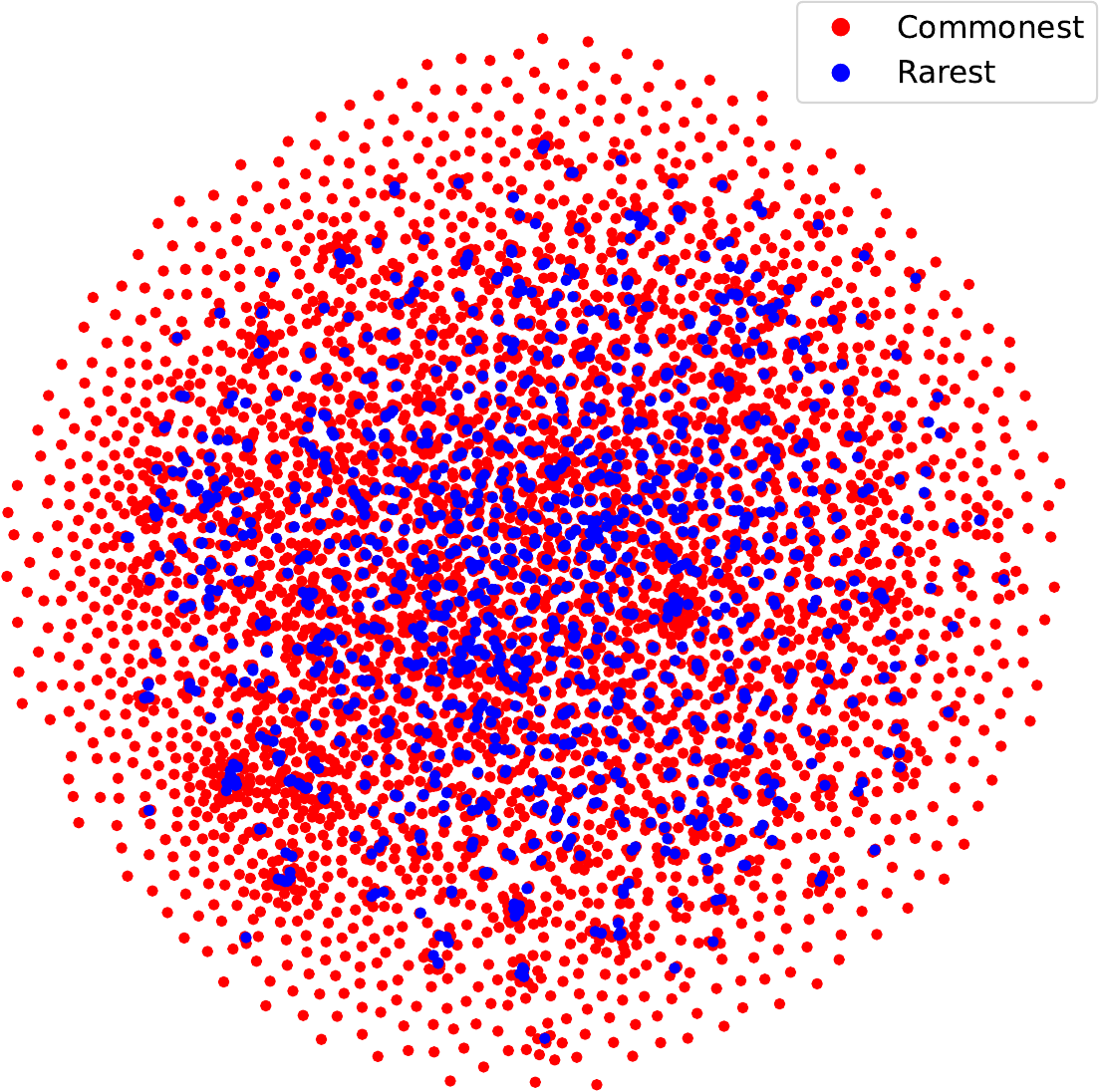}
\label{fig:case:med:after}
\end{minipage}%
}%
\centering
\setlength{\abovecaptionskip}{-0.05cm}   
\setlength{\belowcaptionskip}{-0.1cm}   
\caption{Code Semantics.}
\label{fig:case}
\vspace{-1em}
\end{figure} 

\subsubsection{Case Study}
We visualize  disease representations before and after DRL.
As shown in Figure~\ref{fig:case}, the rare disease prior to DRL exhibits a more random distribution with high entropy, indicating the PCM's struggle to capture their inherent similarities. Because limited EHRs for rare diseases hinder the development of effective CO representations.
In stark contrast, after DRL mapping, rare disease clusters are tightly grouped. This indicates that DRL effectively leverages text knowledge to capture underlying similarities, yielding more meaningful and clinically relevant representations of rare diseases.
Another notable observation is that the distribution difference between rare and common diseases is more pronounced in Diag Pred than in Med Rec. This is because Diag Pred involves more complex relationships due to a higher number of targets, requiring greater changes in DRL. In contrast, Med Rec, with fewer targets, relies more on clearer existing models. This is intuitive, as medications are typically tailored for specific diseases and are relatively straightforward, whereas disease risks are often unpredictable.


\vspace{-0.1cm}
\section{Conclusion}\label{sec:con}
In this paper, we introduce UDC, an innovative framework aimed at enhancing the representation semantics of rare diseases. UDC utilizes discrete representation learning to connect textual knowledge and CO signals, enabling both signals to be in the same semantic space. The framework incorporates condition-aware and task-aware calibration, along with co-teacher distillation tailored for healthcare applications. These advancements significantly enhance the distinguishability and task awareness of encoded representations, as well as the code-level alignment between textual and CO signals.
 Extensive experiments validate the efficacy of our approach. 
However, our model has limitations, including the need to integrate modalities beyond text, which will be explored in future work.


\begin{acks}
xxxxx
\end{acks}

\flushcolsend

\bibliographystyle{ACM-Reference-Format}
\bibliography{main}
\newpage
\clearpage
\appendix

\begin{table}
\centering
\setlength{\abovecaptionskip}{-0.05cm}   
\setlength{\belowcaptionskip}{-0.1cm}   
\caption{Diverse condition encoder. Performance comparison on MIMIC-III Dataset.}
\label{tab:mimic_iii}
\resizebox{0.48\textwidth}{!}{
\begin{tabular}{c|l|ccc} 
\hline
\textbf{Task}              & \textbf{\textbf{Metric}} & \textbf{MLP}               & \textbf{LSTM}              & \textbf{MHA}                                  \\ 
\hline
\multirow{2}{*}{Diag Pred} & Acc@20                   & 0.3324                     & 0.3319                     & \textbf{0.3377}                               \\
                           & Pres@20                  & \multicolumn{1}{l}{0.3632} & \multicolumn{1}{l}{0.3622} & \multicolumn{1}{l}{\textbf{\textbf{0.3713}}}  \\ 
\hline\hline
\multirow{2}{*}{Med Rec}   & Jaccard                  & 0.5252                     & 0.5259                     & \textbf{0.5261}                               \\
                           & F1-score                 & \multicolumn{1}{l}{0.6754} & \multicolumn{1}{l}{0.6750} & \multicolumn{1}{l}{\textbf{\textbf{0.6761}}}  \\
\hline
\end{tabular}}
\end{table}

\section{Diverse Condition Encoder}\label{app:cond}
As evidenced in Table~\ref{tab:mimic_iii}, we find that the choice of condition encoders (MHA) has a minor impact, while Eq.~\ref{eq:7} plays a crucial role. In Eq.~\ref{eq:7}, this normalizing ensures that $\mathbf{f}$'s values fall within a similar range, which helps maintain consistency in the scale of the input features, thereby aiding in training stability and convergence without escalating the model's complexity~\cite{xu2019understanding}.

\section{Diverse Training Methods}\label{app:div}
\begin{figure} 
\centering
\subfigure[Diag Pred]{
\begin{minipage}[t]{0.48\linewidth}
\centering
\includegraphics[width=\linewidth,height=0.75\linewidth]{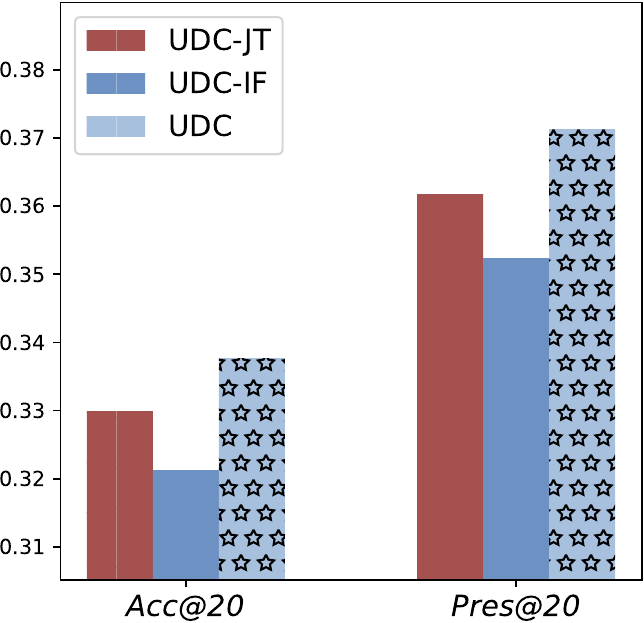}
\label{fig:tm:diag}
\end{minipage}%
}%
\subfigure[Med Rec]{
\begin{minipage}[t]{0.48\linewidth}
\centering
\includegraphics[width=\linewidth,height=0.75\linewidth]{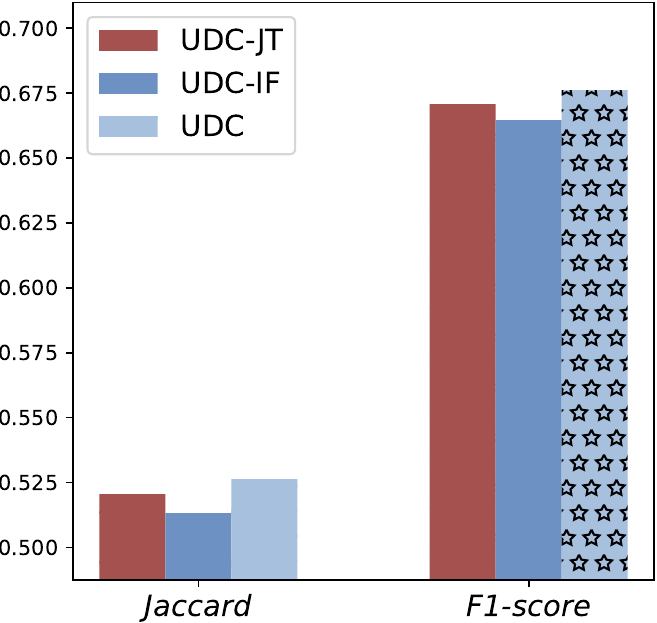}
\label{fig:tm:med}
\end{minipage}%
}%
\centering
\setlength{\abovecaptionskip}{-0.05cm}   
\setlength{\belowcaptionskip}{-0.1cm}   
\caption{Different training methods (MIMIC-III). Please note that UDC refers to the strategy utilized in the manuscript.}
\label{fig:diff}
\vspace{-1em}
\end{figure}

We also experiment with various training methods, such as \underline{j}oint \underline{t}raining ($\theta$, $\Theta$) and \underline{i}nference without \underline{f}ine-tuning, as depicted in Figure~\ref{fig:diff}. Formally,
UDC-JT trains PCM and DRL simultaneously, and we observe that this model initially focuses on learning collaborative signals, leading to DRL training collapse. In contrast, UDC-IF skips fine-tuning and directly performs inference. However, since $\mathcal{F}_{\text{co}}(\cdot)$ does not fully capture the interaction patterns between rare and common diseases, improvements stem primarily from the integration of textual semantic information. From UDC, it is evident that learning these interaction patterns plays a critical role in enhancing the model's overall performance.

\section{Algorithm}\label{app:alg}
The algorithm flow is shown in Algorithm~\ref{alg1}.
\vspace{-1em}
\begin{algorithm}\small 
\caption{The Algorithm of \textit{UDC}} 
\label{alg1} 
\begin{algorithmic}[1] 
\REQUIRE EHR $\mathcal{U}$, Textual Knowledge $\mathrm{T(\cdot)}$, Rare threshold $\eta$;
\ENSURE PCM parameters $\theta$, DRL parameter $\Theta$;
\STATE \textbf{Stage 1: Backbone Training} \Comment{Tuning $\theta$} 
\STATE PCM training $\mathbf{e}_{d}\in \mathbf{E}_{\mathcal{D}}$, $\mathbf{e}_{p} \in \mathbf{E}_{\mathcal{P}}$, $\mathbf{e}_{m} \in \mathbf{E}_{\mathcal{M}}$; 
\STATE PLM Initialization $\mathbf{\Tilde{e}}_{d} \in \mathbf{\Tilde{E}}_{\mathcal{D}}$, $\mathbf{\Tilde{e}}_{p} \in \mathbf{\Tilde{E}}_{\mathcal{P}}$, $\mathbf{\Tilde{e}}_{m} \in \mathbf{\Tilde{E}}_{\mathcal{M}}$;
\STATE \textbf{Stage 2: DRL Training}
\Comment{Frozen $\mathbf{E}$ \& $\mathbf{\Tilde{E}}$, Tuning $\Theta$}
\STATE Split disease into $\mathcal{D}_{\text{com}}$, $\mathcal{D}_{\text{rar}}$ using $\eta$;
\WHILE{not converged}
\STATE Sample disease $d$ from $D_{\text{com}}$;
\STATE Extract PCM \& PLM embedding $\mathbf{e}_{d}$, $\mathbf{\Tilde{e}}_{d}$ in Eq.~\ref{eq:1}-\ref{eq:3};
\STATE Obtain discrete representation $\mathbf{z}_{d}$ and $\mathbf{\Tilde{z}}_{d}$ in Eq.~\ref{eq:5};
\STATE Condition-aware calibration in Eq.~\ref{eq:7};
\STATE Task-aware calibration in Eq.~\ref{eq:10};
\STATE Co-teacher distillation for codebook in Eq.~\ref{eq:12}-\ref{eq:13};
\STATE Optimization in Eq.~\ref{eq:14};
\STATE Update the parameters;
\ENDWHILE
\STATE \textbf{Stage 3: Fine-tuning} 
\Comment{Frozen $\Theta$, Tuning $\theta$}
\STATE Obtain enhanced disease representation $\mathbf{\hat{e}}_{d}$ in Eq.~\ref{eq:15};
\STATE Fine-tuning $\theta$ using Eq.~\ref{eq:2};
\STATE \RETURN{} Parameters $\theta$ \& $\Theta$;
\end{algorithmic}
\end{algorithm}

\begin{table*}[!h]
\centering
\setlength{\abovecaptionskip}{-0.05cm}   
\setlength{\belowcaptionskip}{-0.1cm}   
\caption{Data Statistics across all datasets (Diag Pred || Med Rec). Due to task-specific preprocessing variations, we present data statistics for all tasks. \# means the number of.}
\label{tab:sta}
\resizebox{\textwidth}{!}{
\begin{tabular}{l|ccc||ccc} 
\toprule
\textbf{Items}                & \textbf{MIMIC-III}  & \textbf{MIMIC-IV}     & \textbf{eICU}        & \textbf{\textbf{MIMIC-III}} & \textbf{\textbf{MIMIC-IV}} & \textbf{\textbf{eICU}}  \\ 
\hline
\# of patients / \# of visits     & 6,164 / 9,693              &  26,697 / 99,668             & 8,853 / 10,188  & 35,707 / 44,399      & 46,187 / 154,962     & 114,473 / 124,564        \\
diag. / prod. / med. set size & 4,017 / 1,274 / 192        &   16,906 / 9,026  / 199           & 1,326 / 422 /1,411       & 6,662 / 1,978 / 197 & 19,438 / 10,790  / 200 & 1,670 / 461 / 1,411   \\
avg. \# of visits                      & 1.5725              & 3.7333            & 1.1508    & 1.2434              & 3.3551                & 1.0882            \\
avg. \# of diag per visit             &  27.7807            &   58.2390             & 10.1569  & 17.7373             & 48.9516               & 7.6574              \\
avg.  \# of prod per visit              &  7.7473             &   9.7644             & 32.6515  & 6.1718              & 8.7626                & 27.9025             \\
avg. \# of drug per visit               &  29.6780             &     24.6252          & 15.7981    & 27.1113             & 23.8334               & 17.2664          \\
\bottomrule
\end{tabular}}
\vspace{-1em}
\end{table*}

\section{Dataset Statistics}\label{app:sta}
MIMIC-III is a widely utilized dataset containing EHRs from over 40,000 patients in critical care. MIMIC-IV, the successor to MIMIC-III, expands on this with data from over 70,000 admissions, reflecting more recent practices and broader patient demographics.
eICU comprises health data from over 200,000 patients across various ICU settings in the United States, offering extensive coverage of diverse clinical environments and treatment modalities. We present the dataset statistics after pre-processing~\cite{zhao2024enhancing,yang2023pyhealth} in Table~\ref{tab:sta}.

\section{Further Analysis}\label{sec:4.4}
We conduct additional analyses to gain further insights.

\subsection{Examination of Top-K}\label{sec:4.3.2}
\begin{figure}[!h] 
\vspace{-0.3cm}
\centering
\subfigure[Acc@K]{
\begin{minipage}[t]{0.48\linewidth}
\centering
\includegraphics[width=\linewidth,height=0.7\linewidth]{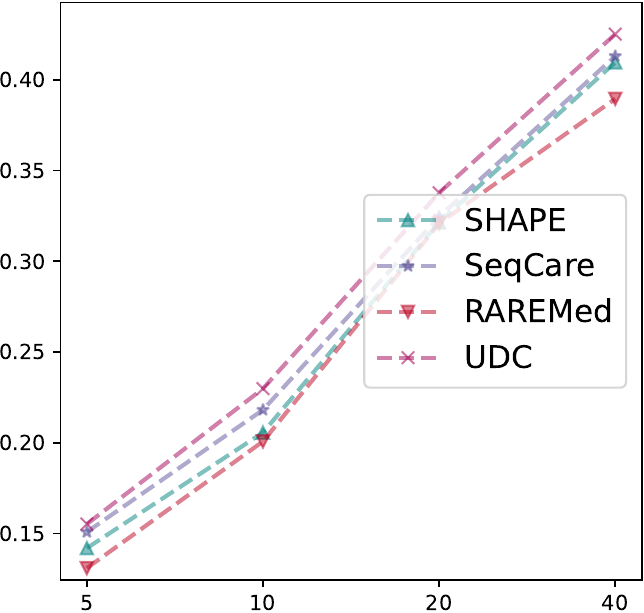}
\label{fig:topk:acc}
\end{minipage}%
}%
\subfigure[Pres@K]{
\begin{minipage}[t]{0.48\linewidth}
\centering
\includegraphics[width=\linewidth,height=0.7\linewidth]{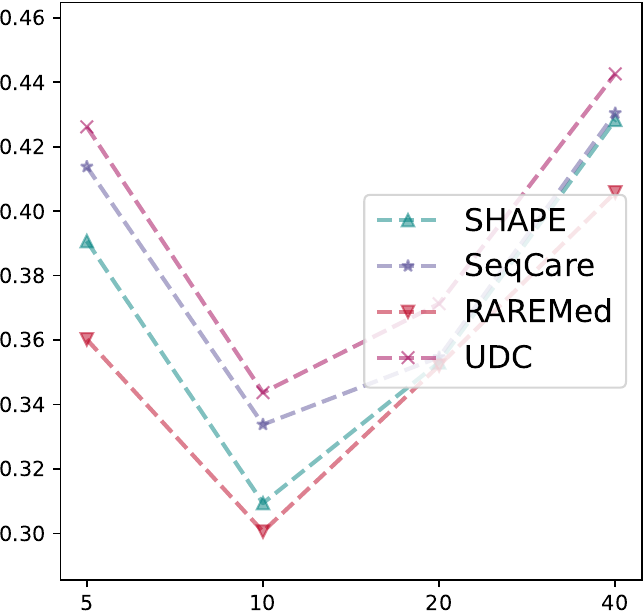}
\label{fig:topk:pre}
\end{minipage}%
}%
\centering
\setlength{\abovecaptionskip}{-0.05cm}   
\setlength{\belowcaptionskip}{-0.1cm}   
\caption{Top-K examination. We test $K=[5,10,20,40]$.}
\label{fig:topk}
\vspace{-0.5cm}
\end{figure}

Top-K evaluation is crucial as it strikes a balance between precise diagnosis and broad screening in Diag Pred~\cite{wang2023stage,xu2023seqcare}.
As shown in Figure~\ref{fig:topk}, all algorithms' Acc@K improve with increasing K, as a larger Top-K captures more relevant medical entities, aiding in the challenging Diag Pred.
Notably, regardless of the specific K value setting, our UDC consistently outperforms the strongest baseline. Moreover, when $K=10$, UDC demonstrates  4\% improvement over the best competing SeqCare. This highlights the effectiveness of our approach in challenging scenarios.
This further demonstrates its broad applicability, a crucial trait for clinical decision support systems, which often require flexibility in the number of diagnoses or treatment options presented.

\subsection{Hyper-parameters Testings}
We further discuss several key hyperparameters.
\begin{figure}[!h] 
\centering
\subfigure[Diag Pred]{
\begin{minipage}[t]{0.48\linewidth}
\centering
\includegraphics[width=\linewidth,height=0.75\linewidth]{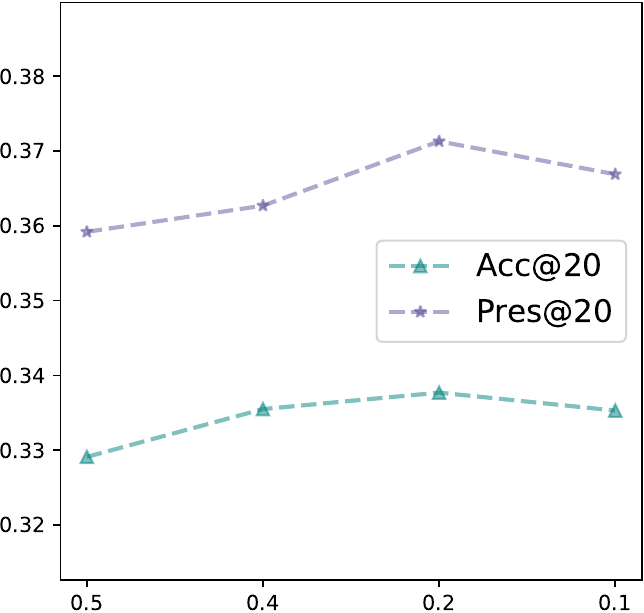}
\label{fig:fur:ratio:dia}
\end{minipage}%
}%
\subfigure[Med Rec]{
\begin{minipage}[t]{0.48\linewidth}
\centering
\includegraphics[width=\linewidth,height=0.75\linewidth]{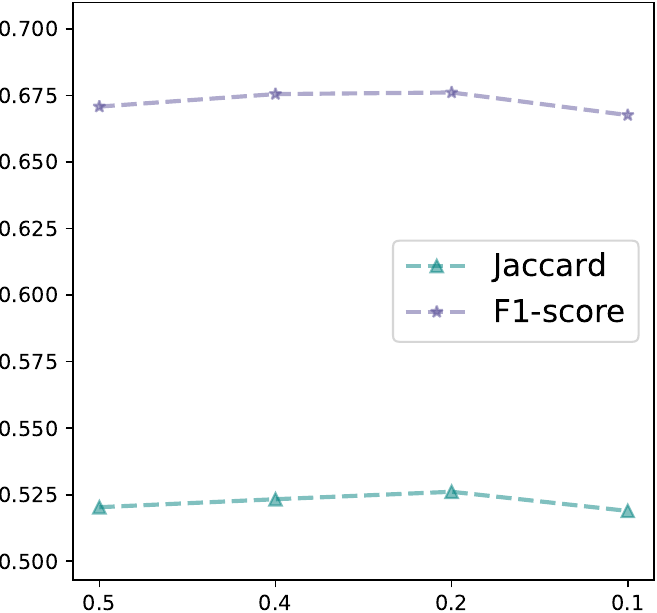}
\label{fig:fur:ratio:med}
\end{minipage}%
}%
\centering
\setlength{\abovecaptionskip}{-0.05cm}   
\setlength{\belowcaptionskip}{-0.1cm}   
\caption{Performance under different ratios. (MIMIC-III)}
\label{fig:fur:ratio}
\vspace{-1em}
\end{figure} 

\noindent\textbf{Rare Ratio $\eta$.}
Our results, as shown in Figure~\ref{fig:fur:ratio}, indicate that UDC achieves the best performance when  $\eta=20\%$. When $\eta$ is too low, the DRL  may not be well-trained from a limited CO-Text pair, making it difficult to obtain semantic alignment between CO and textual spaces. Conversely, with a very high value for $\eta$, UDC does not yield significant improvement. The restricted absolute quantity results in fewer rare disease entity adjustments, exerting minimal influence on the comprehensive sequence representation.
\begin{figure}[!h] 
\centering
\subfigure[Codebook Size]{
\begin{minipage}[t]{0.48\linewidth}
\centering
\includegraphics[width=\linewidth,height=0.75\linewidth]{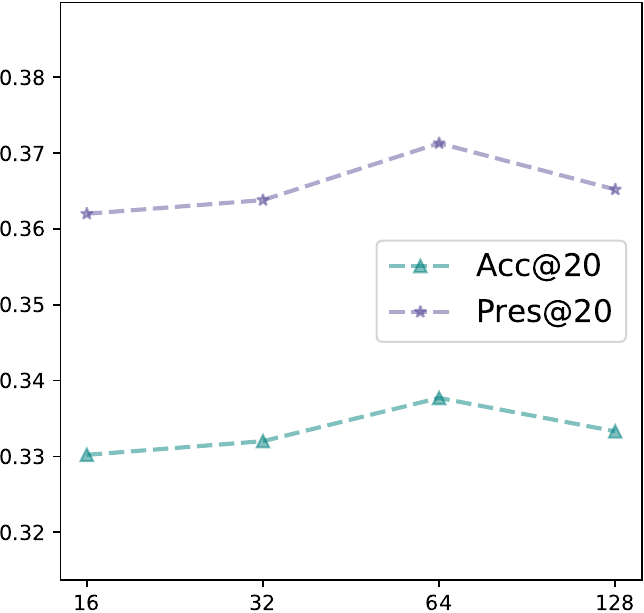}
\label{fig:fur:hyper:code}
\end{minipage}%
}%
\subfigure[Commitment Weight]{
\begin{minipage}[t]{0.48\linewidth}
\centering
\includegraphics[width=\linewidth,height=0.75\linewidth]{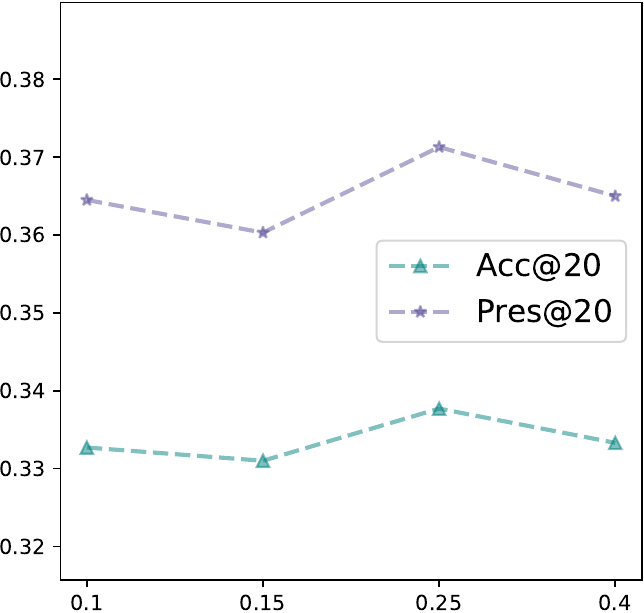}
\label{fig:fur:hyper:aux}
\end{minipage}%
}%
\centering
\setlength{\abovecaptionskip}{-0.05cm}   
\setlength{\belowcaptionskip}{-0.1cm}   
\caption{(a) Performance under different codebook sizes. (b) Performance under different commitment weights. We show the results on MIMIC-III (Diag Pred).}
\label{fig:fur:hyper}
\vspace{-1em}
\end{figure} 

\noindent \textbf{Codebook Size $|\mathcal{C}_{l}|$.}
The codebook size is a critical hyperparameter in the VQ-VAE architecture~\cite{lee2022autoregressive}. A larger codebook size allows the VQ-VAE to capture a richer set of discrete latent features, enabling more detailed and expressive reconstructions of the disease symptoms. However, this comes at the cost of increased computational complexity and potential overfitting, especially when working with limited training data. In contrast, a smaller codebook size can lead to more robust and generalized representations but may struggle to represent the full complexity of the input distribution. Experimentally, we set $|\mathcal{C}_{l}|=64$.

\noindent \textbf{Commitment Weight $\alpha$.}
$\alpha$ is a crucial parameter influencing the quality of absolute code representation and alignment. A larger value enhances the similarity between the encoded representation and the discrete representation, thereby improving cross-domain alignment. However, increasing $\alpha$ may reduce the emphasis on the reconstruction target, potentially leading to negative effects. Experimentally, we set $\alpha=0.25$.

\subsection{Time Complexity Analysis}
Our time cost is competitive with state-of-the-art (SOTA) methods. Specifically, our approach achieves a time complexity of \(O(K \cdot N \cdot D + \frac{|\mathcal{D}| \cdot D^2}{B} + 2 \cdot T^2 \cdot D)\), with an actual runtime of 86 seconds per epoch on the MIMIC-III dataset for medical recommendation tasks using a batch size \(B = 16\). In comparison, SeqCare has a time complexity of \(O(6V \cdot D^2 + 6E \cdot D + 6T^2 \cdot D)\) and runs for 312 seconds per epoch, while GraphCare exhibits a complexity of \(O(T \cdot V \cdot D^2 + T \cdot E \cdot D + T^2 \cdot D + V \cdot D^2)\) with a runtime of 283 seconds per epoch. RAREMed, on the other hand, achieves a time complexity of \(O(2M^2 \cdot D + T^2 \cdot D)\) and runs for 78 seconds per epoch. Here, \(B\) denotes the batch size, \(K\) is the number of codebooks, \(N\) represents the code size per codebook, \(D\) is the embedding size, \(T\) is the sequence length, \(|\mathcal{D}|\) is the number of diseases, \(V\) and \(E\) are the number of graph nodes and edges, respectively, and \(M\) is the pretraining sequence length, which is typically larger than \(T\). Our method demonstrates competitive efficiency while maintaining strong performance.

\subsection{Case Study: Real Prediction}
To intuitively demonstrate the superiority of UDC, we present the medication recommendations for a randomly selected patient.
Specifically, UDC achieves a significantly higher Jaccard  compared to the other baselines. This indicates that UDC can generate diagnostic and treatment suggestions that are much closer to the clinically validated outcomes and better distinguish between positive and negative samples.
Furthermore, F1-score generated by our model is also higher compared to RAREMed. This finding suggests that instead of relying on broad recommendations to enhance performance metrics, our framework offers improved recommendations that effectively balance sensitivity and specificity~\cite{criqui1985sensitivity}.

\begin{table}[!h]\small
\centering
\vspace{-0.3cm}
\setlength{\abovecaptionskip}{-0.05cm}   
\setlength{\belowcaptionskip}{-0.1cm}   
\caption{Example recommendation result.}
\label{tab:medc}
\resizebox{0.48\textwidth}{!}{
\begin{tabular}{cl} 
\toprule
\multicolumn{1}{l}{\textbf{Method}}                                                & \textbf{Recommended Med Set}                                                                                \\
\hline
\begin{tabular}[c]
{@{}c@{}}\textbf{Ground-Truth} \\Num:12 \end{tabular}                         &  \begin{tabular}[c]{@{}l@{}} \textbf{TP:}  [`B05X', `B01A', `A12B', `C07A', `A06A', `C10A', \\ `N02B', `A03B', `C09A', `N06A', `A04A', `C09C']     \end{tabular}                                                                                            \\
\begin{tabular}[c]{@{}c@{}}\textbf{RAREMed}\\Num:12 \\F1-score:0.7500\\Jaccard:0.6000\end{tabular}  & \begin{tabular}[c]{@{}l@{}}\textbf{\textbf{TP:}} [`A03B', `A06A', `A12B', `B01A', `B05X', `C07A', \\ `C09A', `C10A', `N02B'] \\\textbf{\textbf{FN:}} [`A04A', `C09C', `N06A'] \\\textbf{\textbf{FP:}} [`A02B', `A12C', `N02A'] \end{tabular}  \\
\begin{tabular}[c]{@{}c@{}}\textbf{UDC}\\Num: 10 \\F1-score:0.8181\\Jaccard:0.6923\end{tabular}  & \begin{tabular}[c]{@{}l@{}}\textbf{\textbf{TP:}} [`A03B', `A06A', `A12B', `B01A', `B05X', `C07A', \\ `C09A', `C10A', `N02B'] \\\textbf{\textbf{FN:}} [`A04A', `C09C', `N06A'] \\\textbf{\textbf{FP:}} [`A12C']\end{tabular}  \\
\bottomrule
\end{tabular}
}
\end{table}

\end{document}